\documentclass[acmsmall]{acmart}
\usepackage{multirow}
\usepackage{multicol}
\usepackage[most]{tcolorbox}
\usepackage{caption}
\usepackage{subcaption}
\usepackage{enumitem}
\usepackage{threeparttable}
\usepackage{bbding}
\usepackage{algorithm}
\usepackage{algorithmic}

\newcommand\RQOne{(RQ1) Are the interpretations from AIOps models internally consistent?}
\newcommand\RQTwo{(RQ2) Are the interpretations from AIOps models externally consistent?}
\newcommand\RQThree{(RQ3) Are the interpretations from AIOps models consistent across time (i.e., time consistent)?}

\setcopyright{acmcopyright}

\acmJournal{TOSEM}
\acmVolume{}
\acmNumber{}
\acmArticle{}
\acmMonth{8}

\definecolor{custom-gray}{cmyk}{0, 0, 0, 0.7, 1.00}
\newtcbtheorem[no counter]{Summary}{\hskip-0.97em}{enhanced,drop shadow={black!50!white},
  coltitle=white,
  top=0.15in,
  attach boxed title to top left=
  {xshift=1.5em,yshift=-\tcboxedtitleheight/2},
  boxed title style={size=small,colback=custom-gray}
}{summary}

\makeatletter
\newcommand{\printfnsymbol}[1]{%
  \textsuperscript{\@fnsymbol{#1}}%
}
\makeatother

\begin{document}


\title{Towards a consistent interpretation of AIOps models}

\author{Yingzhe Lyu}
\authornote{These authors contributed equally to the work.}

\email{ylyu@cs.queensu.ca}
\affiliation{%
  \institution{Software Analysis and Intelligence Lab (SAIL), Queen's university}
  \city{Kingston}
  \state{ON}
  \country{Canada}
}

\author{Gopi Krishnan Rajbahadur\printfnsymbol{1}}
\email{gopi.krishnan.rajbahadur1@huawei.com}
\affiliation{%
  \institution{Centre for Software Excellence, Huawei Canada}
  \state{ON}
  \country{Canada}}

\author{Dayi Lin\printfnsymbol{1}}
\email{dayi.lin@huawei.com}
\affiliation{%
  \institution{Centre for Software Excellence, Huawei Canada}
  \state{ON}
  \country{Canada}}

\author{Boyuan Chen\printfnsymbol{1}}
\email{boyuan.chen1@huawei.com}
\affiliation{%
  \institution{Centre for Software Excellence, Huawei Canada}
  \state{ON}
  \country{Canada}}

\author{Zhen Ming (Jack) Jiang}
\email{zmjiang@cse.yorku.ca}
\affiliation{%
  \institution{Lassonde School of Engineering, York University}
  \state{ON}
  \country{Canada}}


\begin{abstract}
Artificial Intelligence for IT Operations (AIOps) has been adopted in organizations in various tasks, including interpreting models to identify indicators of service failures. To avoid misleading practitioners, AIOps model interpretations should be consistent (i.e., different AIOps models on the same task agree with one another on feature importance). However, many AIOps studies violate established practices in the machine learning community when deriving interpretations, such as interpreting models with suboptimal performance, though the impact of such violations on the interpretation consistency has not been studied.

In this paper, we investigate the consistency of AIOps model interpretation along three dimensions: internal consistency, external consistency, and time consistency. We conduct a case study on two AIOps tasks: predicting Google cluster job failures, and Backblaze hard drive failures. We find that the randomness from learners, hyperparameter tuning, and data sampling should be controlled to generate consistent interpretations. AIOps models with AUCs greater than 0.75 yield more consistent interpretation compared to low-performing models. Finally, AIOps models that are constructed with the Sliding Window or Full History approaches have the most consistent interpretation with the trends presented in the entire datasets. Our study provides valuable guidelines for practitioners to derive consistent AIOps model interpretation.
\end{abstract}



\maketitle

\section{Introduction}
\label{sec:introduction}

Ensuring the quality of service of cloud computing platforms is extremely pivotal. A recent IDG survey~\cite{cloud_2020} reports that 92\% of organizations leverage cloud computing platforms for running their applications. Therefore, it is extremely important to ensure that these cloud computing platforms remain highly available and efficient, particularly since failures in cloud computing platforms are estimated to cost up to \$700 billion annually~\cite{merker_2020}. For instance, a recent survey~\cite{alsop_2020} pegs the average cost per hour of an organization's server downtime to be anywhere between \$300,001 to \$400,000.  

Cloud computing platforms generate a tremendous amount of data which is impossible to be analyzed manually. Recently, it has become increasingly common for organizations to use AIOps (Artificial Intelligence for IT Operations) to leverage such generated data to ensure the quality of service and high availability of cloud computing platforms ~\cite{LiTOSEM20,ZhaoFSE20,BansalSEIP20,BanerjeeOpML20,NedelkoskiCCGRID19,DangICSE19}. AIOps leverages machine learning learners to construct machine learning models (hereafter AIOps models) with operations data collected from the cloud computing platforms (e.g., logs and alert signals) to enable quality assurance tasks such as predicting hard drive failures~\cite{LiTOSEM20}, job termination~\cite{SayedICDCS17}, service outages~\cite{ZhaoFSE20} and performance issues~\cite{LimICDM14}. Note that we use the term ``learner” to refer to a machine learning algorithm (e.g., Random Forest) and the term ``model” to refer to a trained machine learning model (e.g., a Random Forest model trained on disk failure data). 


In addition to using AIOps models to predict failures and outages on a cloud computing platform, several prior studies also interpret AIOps models to identify association between different factors and occurrences of certain failures or outages to make operational and business decisions. 
For instance,~\citet{ChenASE20} interpret their deep learning based incident prediction model and find that incident-occurring environment features are one of the most important indicators of a potential incident happening in their platform. Similarly,~\citet{ZhaoFSE20} interpret their XGBoost based incident prediction model and find that in their systems, database related issues are the root cause for the incidents. ~\citet{LiTOSEM20} interpret their AIOps models to refine their automated alert management system.

These derived interpretations of AIOps models should be consistent (i.e., the feature importance ranking from different models’ interpretations agree with one another), to avoid misleading practitioners. For instance, as we mentioned earlier,~\citet{ChenASE20} derive interpretations from their deep learning based incident prediction model. However, if their incident prediction model is retrained (e.g., on a local instance of the same training data) and it produces a different set of features being the most important feature associated with incidents, the practitioners might not know which interpretation to trust. Such a case is indeed possible as~\citet{PhamASE20} show, when a model is retrained, even on the same training dataset, its performance might change (sometimes even drastically). Thus, it is possible that the interpretations that are derived from the retrained incident prediction model might also vary. More generally, recent studies~\cite{tantithamthavornTSE2018,tantithamthavorn2018TSE2} in several domains point out that many factors (e.g., different hyperparameters) could impact the consistency of derived interpretations of a machine learning model. 

However, to the best of our knowledge, none of the prior studies in the field of AIOps investigates the factors that could impact the consistency of the derived interpretations of AIOps models. Though the factors relating to the consistency of derived interpretations have been explored by a handful of studies~\cite{jiarpakdee2020TSE,rajbahadurTSE2020,tantithamthavorn2018TSE2,tantithamthavornTSE2018} in other domains, it is pivotal to explore them in the context of AIOps for the following reasons:

\begin{itemize}
\item \citet{chenFSE2020} and~\citet{DangICSE19} point out there are several challenges that are unique to the field of AIOps. These unique challenges might present unique factors (which are typically not prevalent in other domains such as machine learning) that impact the consistency of the interpretations derived from AIOps models differently. For instance, as~\citet{DangICSE19} explain, AIOps models are constantly updated/retrained to keep up with the rapidly evolving data. Such rapid retraining of AIOps models could mean that the derived interpretations of an AIOps model could change with every retraining. However, as we explain earlier, consistency of derived interpretation is pivotal for practitioners. Therefore, it is important to understand how the factors that are unique to AIOps impact the consistency of interpretations derived from AIOps.

\item The operations data that is typically used to build AIOps models is very different from the data used in other domains like machine learning. Operations data is typically a mixture of temporal (e.g., log data) and spatial (e.g., hardware configurations) data. In addition, it may also contain a mixture of heterogeneous data types like numeric, ordinal and nominal values, making it starkly different from data used in other machine learning tasks (which is typically either spatial or temporal in nature).

\item As~\citet{menziesSOFTWARE2019} and~\citet{rayICSE2016} show, findings from other domains like machine learning do not necessarily generalize in software analytics domains like AIOps. For example, both~\citet{menziesSOFTWARE2019} and~\citet{rayICSE2016} show that language models used in the machine learning domain typically yield spurious results on software engineering related data. Furthermore,~\citet{menziesSOFTWARE2019} warns us that simply using the methods and findings outlined in the field of machine learning on software engineering data might yield suboptimal results.

\end{itemize}

Therefore, in our study, we aim to better understand the factors that could impact the consistency and the practical adoption of the derived interpretations of an AIOps model. In our paper, we first propose a set of rigorous criteria to thoroughly assess the consistency of the interpretations derived from AIOps models. We do so through a case study on two publicly available operations datasets (the Google cluster trace dataset~\cite{reissGOOGLE2011} and the BackBlaze hard drive statistics dataset~\cite{backblazeDATA}) that have been widely used by many prior studies in AIOps~\cite{BotezatuKDD16,SayedICDCS17,RosaIWQOS15}. In particular, our proposed criteria investigate how different factors impact the AIOps model interpretation along three key dimensions: \textbf{Internal consistency}, \textbf{External consistency}, and \textbf{Time consistency}.

\begin{itemize}

\item \smallskip\noindent\textbf{Internal consistency}~\cite{PhamASE20} (a.k.a., model reproducibility) captures the similarity between the derived interpretations of an AIOps model trained from the same setup (i.e., same training data and same implementation) across multiple executions. In other words, internal consistency checks if the interpretations derived from an AIOps model is reproducible. For instance, as~\citet{PhamASE20} show, and a plethora of prior studies~\cite{Hutson725,GundersenAAAI18,PhamASE20,pineau2020improving} warn, unless randomness involved during  the training process of machine learning models is controlled, the results might not be reproducible. As a result, the derived interpretations might not be internally consistent which would make it impossible for the managers and DevOps engineers to choose which interpretation to act on. Despite that, many of the AIOps studies do not take specific steps to ensure reproducibility. For instance, both~\citet{ZhaoFSE20} and~\citet{LiTOSEM20} use deep learning models to construct their AIOps models. However, neither of those studies take explicit methods to control the randomness involved which could in turn, impact the reproducibility of the interpretation derived from these AIOps models. Therefore in RQ1 (Are the interpretations from AIOps models internally consistent?), we study the impact of potential sources of randomness during the model training, which can impact the internal consistency of the derived interpretations of the AIOps models.

\smallskip\noindent\textbf{Results.} All the three studied sources of randomness (i.e., inherent randomness from learners, randomized hyperparameter tuning and data sampling randomness) impact the consistency of AIOps model interpretation. Sampling randomness introduces the largest scale of inconsistency to AIOps model interpretation. When all the three sources are controlled during the training of an AIOps model, the derived interpretation of an AIOps model is internally consistent.


\item \smallskip\noindent\textbf{External consistency}~\cite{rajbahadurTSE2020} captures the similarity between the derived interpretations of similar-performing AIOps models on a given dataset. In other words, external consistency sanity checks if the models that have the similar performance report similar interpretations for a dataset. Typically, interpretable models are preferred by the practitioners to derive interpretations. Several recent studies in AIOps~\cite{LiTOSEM20} and machine learning~\cite{rudin2019stop} argue that interpretable models even with lower performance are preferred in the AIOps context when high-performing models are not easily interpretable.~\citet{rajbahadurTSE2020} recently show that the derived interpretation between different machine learning models could vary considerably. In addition,~\citet{liptonQUEUE2018} and~\citet{molnar2019} warn that such inconsistency could be exacerbated if low-performing models are used to derive interpretations. Intuitively, interpretations derived from a low-performing interpretable model could be trustworthy only if the interpretable model has the same interpretation as other machine learning models on a given dataset (at least among the similar-performing models). In other words, different models at the same performance level should generate similar interpretations. However, there are no studies which examine the relation between the model performance and model interpretations. In particular, it is not clear if using high-performing models would yield more consistent interpretations than using low-performing models. Hence, in this paper,  we assess and compare the external consistency of the  interpretations from models at different performance levels (i.e., similarities of model interpretations among low-performing models vs. similarities of model interpretations among high-performing models). Such study is even more important for low-performing models since several prior studies question if low-performing models might generate inconsistent interpretations~\cite{jiarpakdeeEMSE2020,molnar2019,liptonQUEUE2018}. Otherwise, interchangeably using models to derive interpretations might be misleading and might result in misguided decisions. Therefore in RQ2 (Are the interpretations from AIOps models externally consistent?), we investigate the external consistency of derived interpretations of AIOps models at different performance levels.

\smallskip\noindent\textbf{Results.} The interpretations derived from high-performing AIOps models (with a minimum acceptable AUC of 0.75) are more consistent than those from low-performing AIOps models. The interpretations derived from high-performing AIOps models exhibit strong external consistency.

\item \smallskip\noindent\textbf{Time consistency}~\cite{LiTOSEM20} captures the similarity between the derived interpretations of an AIOps model across different time periods (i.e., as the AIOps models evolve). In other words, time consistency checks if the interpretation derived from an AIOps model remains generalizable across time. Interpretations derived from AIOps models may be used to make business decisions and process optimizations that have a long-lasting effect. Hence, it is pivotal that the interpretation of an AIOps model should not only just reflect the trend from the most recent data on which it was trained, but also should capture and reflect the trends observed over a longer period of time. However, previous works in defect prediction~\cite{bangash2020time} and AIOps~\cite{LiTOSEM20} show machine learning models trained on one time period do not generalize well when tested on a different time period and their derived interpretations could also vary depending on the size of the training data. Therefore, in RQ3 (Are the interpretations from AIOps models consistent across time (i.e., time consistent)?), we examine which of the commonly used model updating approaches allows the interpretations of the updated models to be consistent over long periods of time. 

\smallskip\noindent\textbf{Results.} The derived interpretations from AIOps models constructed with the Sliding Window and Full History approaches best capture the trends present across time periods in the entire dataset and exhibit strong time consistency.

\end{itemize}

The main contributions of our paper are the following:

\begin{enumerate}
    \item This is the first work that studies the factors that impact the consistency of AIOps model interpretations. 
    \item We propose a set of rigorous criteria that enables researchers and practitioners to assess the consistency of their derived interpretations from AIOps models.
    \item We provide several actionable guidelines to improve the consistency of interpretations derived from AIOps models.
    \item To foster replicability of our findings and promote open science, we make the datasets and the code to conduct our study publicly available.
 
\end{enumerate}

\smallskip\noindent\textit{Paper organization.} The rest of the paper is organized as follows: in Section~\ref{sec:motivation} we present a motivational example that motivates the need for our study in AIOps. In Section~\ref{sec:background} we provide the background and related work of AIOps, and introduce the research questions that we investigate in our paper. In Section~\ref{sec:case_study_setup} we explain our case study setup. Section~\ref{sec:rq1}, Section~\ref{sec:rq2}, and Section~\ref{sec:rq3} present each of our RQs. Section~\ref{sec:discussion} discusses the results, limitations of the results and potential future areas for investigation. 
In Section~\ref{sec:guideline}, based on our findings from the results of the studied RQs, we outline several practical guidelines for AIOps researchers and practitioners. In Section~\ref{sec:threats} we discuss the threats to validity of our study. Finally, in Section~\ref{sec:conclusion} we conclude our study. 
\section{A Motivational Example}
\label{sec:motivation}

Lei is a DevOps engineer. They are responsible for monitoring and maintaining a batch processing job that runs daily. The batch processing job runs on a cluster and can take hours to finish. In the past, they have noticed that the job may randomly fail and they would need to re-submit the job for execution. Recently the analytics team helped Lei build an AIOps model, which uses the traces collected from the cluster to predict if the job is going to fail in the next half hour. This model is selected from a pool of best-performing candidate models by the analytics team after considering various aspects such as performance and training costs. The analytics team advised Lei to use the AIOps model in two ways:
\begin{enumerate}
    \item \textbf{Predictive}: When the model predicts the job is going to fail in the next half hour with high confidence, they can manually terminate and re-submit the job, to save execution time.
    \item \textbf{Explanatory}: By interpreting the trained model, the analytics team advised them that the model learnt that one of the most impactful factors that is associated with the job failure is the launch of a specific competing job from another team. Therefore from an operations perspective they should contact the other team to schedule the competing job at a different time of the day.
\end{enumerate}

\textbf{Scenario 1}: To productionize the model, the analytics team provided Lei with the data and code used to train the model. Lei executed the provided training code on the same data, only to realize that the model they trained provided a different interpretation compared to that of the model trained by the analytics team. In particular, the model they trained does not consider the competing job as an important factor. Lei is confused about which interpretation to trust, and whether they should reschedule the competing job with the other team. In this scenario, Lei is concerned about the internal consistency of the AIOps model.

\textbf{Scenario 2}: To avoid performance drift of the model in production, the model needs to be periodically retrained with latest data. However the current model provided by the analytics team is too costly to retrain. Lei asked the analytics team if there is any alternative candidate model that is cheaper to retrain. The analytics team sent over a new, lighter model, which is cheaper to retrain albeit at a slightly reduced performance. However, Lei realized that even though the new model was trained on the same data as the current one, it provides a different interpretation, and does not consider the competing job as an important factor, posing a dilemma for Lei about whether to reschedule the competing job with the other team. In this scenario, Lei is concerned about the external consistency of the AIOps model.

\textbf{Scenario 3}: The model was deployed in production and was scheduled to be retrained with new data everyday at midnight. Lei contacted the other team about rescheduling the competing job, as advised by the analytics team based on the interpretation of the deployed model. Rescheduling jobs may lead to a lot of downstream administrative and operative changes, and therefore is expensive to communicate and perform. However, in just a few days, before the rescheduling was scheduled to take effect, Lei noticed that the recently updated model no longer considers the competing job as an important factor. Such drastic changes in the model interpretation led to a lot of confusion among teams, and wasted effort in rescheduling jobs. In this scenario, Lei is concerned about the time consistency of the AIOps model.
\section{Background}
\label{sec:background}

In this section, we first describe the existing work on AIOps. Then we present the current practices of deriving AIOps interpretation. Finally, we present our criteria for assessing the consistency of AIOps interpretation.

\subsection{Existing work on AIOps}

We first introduce the common AIOps applications. Then we discuss the existing work on AIOps interpretation. Finally, we present the reproducibility concerns in general machine learning tasks.

\subsubsection{AIOps Applications}
Although huge efforts have been devoted to cloud computing systems for ensuring the quality of services, various types of issues (e.g., job termination, hard drive failure, and performance anomalies) are unavoidable.
To ensure the reliability of online services, we must resolve and manage these issues in a timely manner, as failing to do so might cause unavailable services and huge financial losses.
AIOps solutions contribute to issue management in two phases: (1) AIOps solutions aim to predict whether certain issues would occur by learning from the historical data; and (2) after the issues occur, AIOps solutions aim to help mitigate the issues (e.g., automated problem diagnosis) or provide suggestions to the domain experts (e.g., incident triage).

\noindent\textbf{Issue Prediction.} 
Many of the prior works focus on analyzing monitoring data for predicting the occurrence of various types of issues~\cite{BotezatuKDD16,ChenWWW19,SayedICDCS17,LinFSE18,LiTOSEM20,MahdisoltaniATC17,RosaIWQOS15,XuATC18,HeFSE18,LimICDM14,ZhaoFSE20,LiDSN14,LiRESS17}.

Lin et al.~\cite{LinFSE18} and Li et al.~\cite{LiTOSEM20} predict node failures in large-scale cloud computing platforms by building machine learning models from temporal (e.g., CPU utilization metrics), spatial (e.g., location of a node), and config data (e.g., build data). Similarly, Li et al~\cite{LiDSN14,LiRESS17} build tree-based models to predict hard drive failures. Botezaku et al.~\cite{BotezatuKDD16}, Mahdisoltani et al.~\cite{MahdisoltaniATC17}, and Xu et al.~\cite{XuATC18} leverage SMART-based analysis to build a machine learning pipeline for predicting hard drive failures in large-scale cloud computing platforms. Chen et al.~\cite{ChenWWW19} collect and analyze the alert data and its dependencies to predict outages in the whole cloud systems. Zhao et al.\cite{ZhaoFSE20} propose a deep learning-based approach, \textit{eWarn}, which leverages textual (e.g., keywords in incident tickets) and statistical features (e.g, alert count) to predict incident occurrences. El-Sayed et al.~\cite{SayedICDCS17} and Rosa et al.~\cite{RosaIWQOS15} predict job failures from trace data collected from Google cloud computing platform. Lim et al.~\cite{LimICDM14} leverage performance metrics to cluster performance issues into the recurrent and unknown ones.


\noindent\textbf{Issue Mitigation.}
Issues in online services need to be mitigated in a timely manner. Many of the prior works focus on triaging~\cite{ChenASE19,ChenASE20,BansalSEIP20}, diagnosing~\cite{ZhangDSN05,LuoKDD14,BanerjeeOpML20,JehangiriICCC14,ChenFSE20}, and managing issues~\cite{LouASE13,LouASE17,XueDSN16,XueTNSM18,JiangFSE20,LinICSE16}, which benefit the mitigation process. 

\noindent\textit{Triaging.} Chen et al.~\cite{ChenASE19} propose a deep learning-based technique to improve the current incident triage process (e.g., distributing the new incident to the responsible team). Chen et al.~\cite{ChenASE20} perform an empirical study on characterizing incidents in online systems and propose DeepIP, a technique to detect incidental incidents (i.e., incidents that are less severe and last for a short period of time), which can reduce the incident triage efforts. Bansal et al.~\cite{BansalSEIP20} propose DeCaf, a Random Forest-based framework to correlate telemetry data with performance regressions. In addition, the detected performance regressions are automatically triaged to on-site engineering team. 

\noindent\textit{Diagnosing.} Zhang et al. ~\cite{ZhangDSN05} propose an ensemble of models to automatically diagnose performance problems. Chen et al.~\cite{ChenFSE20} propose LiDAR, a deep learning-based approach to linking similar incidents based on historical information.  Luo et al.~\cite{LuoKDD14} mine time-series data and event data to discover correlations between them, which could improve the incident diagnosis process. Banerjee et al.~\cite{BanerjeeOpML20} discuss challenges in performance diagnosis in a hybrid-cloud enterprise software environment. Jehangiri et al.~\cite{JehangiriICCC14} present techniques to diagnose performance anomalies using time-series datasets.

\noindent\textit{Managing.} Jiang et al.~\cite{JiangFSE20} analyze the similarity between incident descriptions and their corresponding troubleshooting guide to facilitate incident management. Lou et al.~\cite{LouASE13,LouASE17} develop a software analytic-based system to resolve scalability, reliability, and maintainability of data-driven incident management systems. Xue et al.~\cite{XueDSN16,XueTNSM18} proactively reduce performance tickets by predicting usage series in cloud data centers. Lin et al.~\cite{LinICSE16} propose a data mining-based technique to detect emerging issues (a sudden burst of new issues) by analyzing historical issues.

\subsubsection{AIOps Interpretation}

The importance of interpretability of machine learning-based systems has drawn increasing attention recently as it is related to the trustworthiness, reliability, and quality of the systems. In the context of AIOps, it is essential that we can reason about the model recommendations to make business decisions (e.g., replacing failing hard drives) and improve the status quo (e.g., improving monitoring infrastructure). Here we discuss prior studies on interpretations of AIOps solutions.

Li et al.~\cite{LiTOSEM20} study the importance scores of a random forest-based model and find that alert data is the most important feature set for node failure prediction. 
Zhao et al.~\cite{ZhaoFSE20} leverage a model-agnostic technique, LIME~\cite{RibeiroKDD16}, to generate interpretable report for incident prediction. This technique is mainly used for interpreting each individual prediction.
Bansal et al.~\cite{BansalSEIP20} propose a technique to extract ranked list of rules from random forest-based models, which can be used to explain predicted performance regression. 
Li et al.~\cite{LiDSN14,LiRESS17} present that by using interpretable models (e.g., regression tree-based models), users can derive more meaningful decisions in reducing the hard drive failure rate, which is preferred over ANN-based models in such context.
Chen et al.~\cite{ChenASE19} and Jiang et al.~\cite{JiangFSE20} all leverage deep learning-based systems for mitigating incidents. The deep learning-based systems contain multiple components for reducing the data noise and adjusting loss functions. They interpret the importance of these components by comparing the performance before and after enabling these services.



\subsubsection{Reproducibility of Machine learning}

Recent studies~\cite{Hutson725,GundersenAAAI18} have shown that reproducibility has drawn increasing attention in the machine learning field in recent years. Many research papers did not include sufficient information (e.g., dataset and experiment code), which makes it difficult for other researchers to reproduce the experiment results. Furthermore, even with the same experiment setup, there is still variance in the outcomes due to the stochastic nature of machine learning applications~\cite{PhamASE20}. 
Hence, huge efforts have been devoted to evaluating and improving the reproducibility in the machine learning field. Pham et al.~\cite{PhamASE20} study the variance of deep learning applications and find that non-implementation level factors can cause the accuracy and training time to fluctuate. Pineau et al.~\cite{pineau2020improving} propose a reproducibility checklist before submitting papers to NeurIPS, to improve the quality of scientific contributions. Various guidelines (e.g., Model Cards~\cite{modelcards} and Datasheets~\cite{datasheets}) have also been proposed to help practitioners and researchers to improve the reproducibility of their machine learning applications. 

The interpretation of machine learning applications also suffers from the reproducibility issues. Fan et al.~\cite{FanTIFS21} assess the quality of five interpretation techniques in Android malware analysis applications, where they evaluate the stability, robustness, and effectivenss of the interpretations. Warnecke et al.~\cite{WarneckeESSP20} also study the similar dimensions of interpretation in security domain. They both find that different interpretation techniques could generate different interpretation results for the same prediction result. Other studies~\cite{AdebayoNIPS18,Camburuarxiv19,YehNIPS19} also indicate that the interpretation results cannot be trusted when the results cannot be reproduced.

Our study is different from the previous work in three aspects: (1) our study focuses on the consistency of AIOps interpretation, in particular, we study the internal, external, and time consistency by leveraging one interpretation technique; (2) we leverage the model-level interpretation technique, i.e. the feature importance ranking, to extract the general trend and knowledge in the AIOps model, where previous studies mostly focus on instance-level interpretation; and (3) our work is conducted in the AIOps domain with constantly evolving data, which requires the interpretation to be able to capture the general trends across time.

\subsection{The current practices of deriving interpretations from AIOps models}
We first introduce the the overall process of deriving interpretations from AIOps models. Then we discuss the potential issues with recent AIOps studies.

\subsubsection{The overall process}
As we discussed in the previous section, AIOps solutions have been used extensively for a variety of tasks (e.g., incident prediction~\cite{LiTOSEM20,ZhaoFSE20}, performance analysis~\cite{BansalSEIP20,BanerjeeOpML20}, anomaly detection~\cite{NedelkoskiCCGRID19}, and business decision making~\cite{DangICSE19}).
The quality of AIOps solutions is evaluated from various dimensions such as performance,  scalability, and maintainability. Although many studies leverage the interpretations of AIOps models to support critical decisions, few studies have been done to systematically assess the quality of interpretations of AIOps models. 

\begin{figure}
    \centering
    \newlength{\imagewidth}
    \settowidth{\imagewidth}{\includegraphics{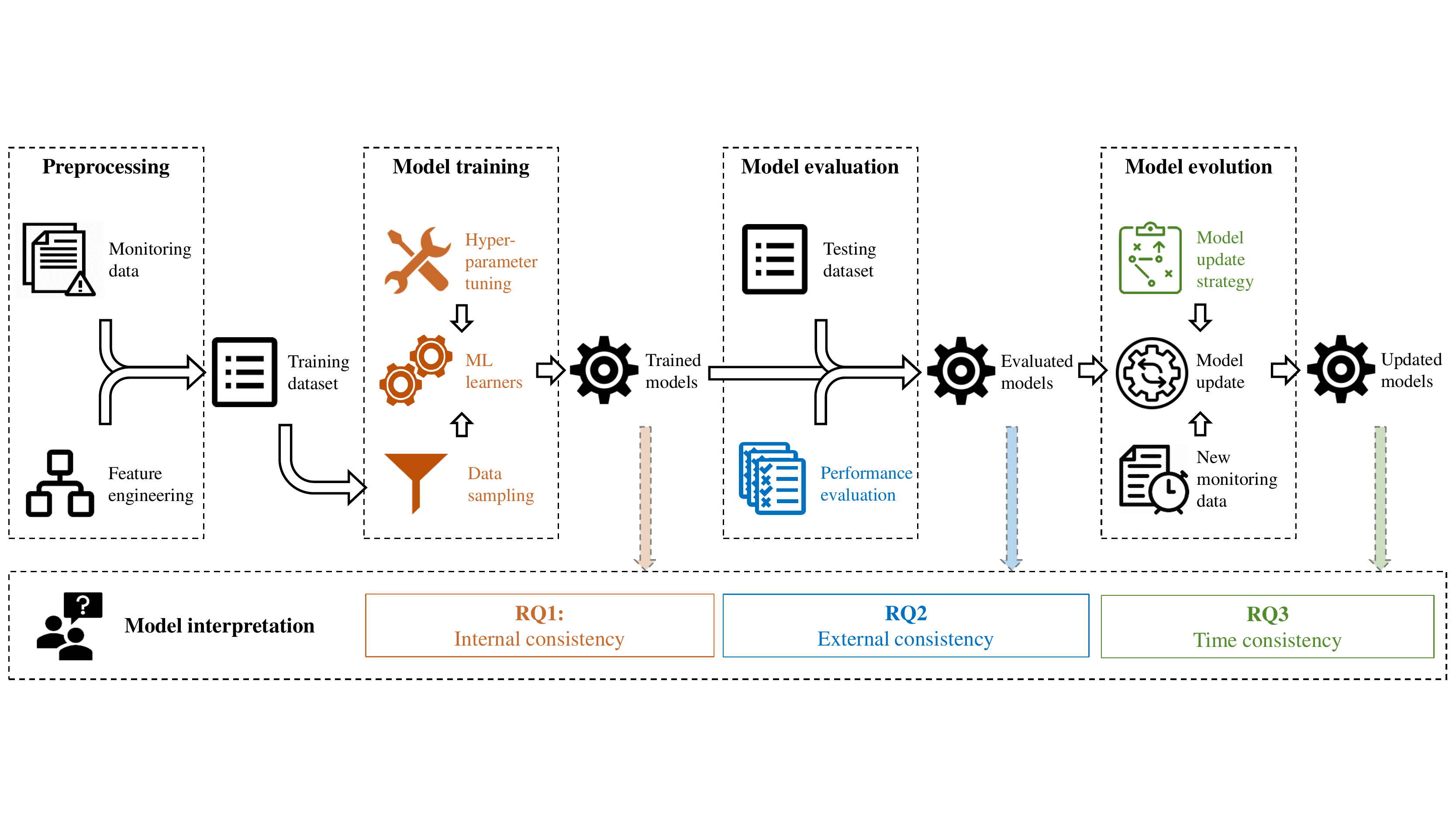}}
    \includegraphics[trim=0 0.09\imagewidth{} 0 0.1\imagewidth{}, clip, width=\textwidth]{pics/Background_process.pdf}
    \caption{An overall process of deriving interpretations from AIOps models, our criteria for assessing the consistency of AIOps model interpretation is highlighted along the process.}
    \label{fig:background_pipeline}
\end{figure}

Figure~\ref{fig:background_pipeline} shows an overall process of deriving interpretations from AIOps models. We divide the pipeline into four major phases:

\begin{enumerate}
    \item The \emph{Preprocessing} phase refers to the process of continuously collecting monitoring data and transforming it into readily available features as inputs for ML models. This is a common practice adopted in many AIOps solutions~\cite{LyuTOSEM21,LiTOSEM20}. In this phase, the training dataset is generated. 
    
    \item The \emph{Model Training} phase refers to the process of training ML models using preprocessed features. It consists of three steps: data sampling, ML learner fitting, and hyperparameter tuning. Data sampling is for selecting a representative subset of a large dataset. ML learner fitting is to select the appropriate learner for the task and fit a model with the learner. Hyperparameter tuning is to find the optimal parameters for the models. In this phase, the trained models are generated.
    
    \item The \emph{Model Evaluation} phase refers to evaluating the performance of the trained models using the testing dataset. The testing dataset is not seen by the model during the model training phase to prevent data leakage. In this phase, the evaluated models are generated. 
    
    \item The \emph{Model Evolution} phase refers to the practice of deploying the models and constantly updating models. Various model update strategies are applied to reflect the trends contained in the newly available period of monitoring data and mitigate the impact of concept drift~\cite{GamaSBIA04,HarelICML14}. In this phase, the updated models are generated.
\end{enumerate}

The overall process is iteratively conducted until the model performance is above a certain threshold. The interpretation can be derived from the trained models, evaluated models, and the updated models. 
There are two general approaches to interpret machine learning models: the model-specific approach and the model-agnostic approach. The model-specific approach is mainly used with models that are intrinsically interpretable. Examples of intrinsically interpretable models include linear regression models and decision tree-based models. On the other hand, some machine learning models are difficult to interpret due to their complex internal structures (e.g., deep neural networks). To interpret these models, practitioners mainly use the model-agnostic approach which explains the models in a post-hoc manner. There are many model-agnostic techniques in the existing literature, such as LIME~\cite{RibeiroKDD16} and permutation feature importance.
There are two types of interpretation results that could be produced: model-level interpretation and instance-level interpretation. The model-level interpretation is to understand how the models work internally, while the instance-level interpretation is to explain each single prediction.
In this paper, we use the permutation feature importance, which is a model-level, model-agnostic interpretation approach, to interpret all the studied AIOps models. 
We explain the approach and rationale in detail in Section~\ref{sec:case_study_setup}.

\subsubsection{Potential issues with recent AIOps studies}
\label{sec:backgroud:issues}

\begin{table}
\centering
\caption{The potential neglected key dimensions of interpretation consistency.}
\label{tab:background_issues}
\begin{tabular}{lccc}
\toprule
Paper & Internal Consistency & External Consistency & Time Consistency \\
\midrule
    P1~\cite{LiTOSEM20} & \XSolid & \XSolid & \XSolid \\
    P2~\cite{ZhaoFSE20} & \XSolid & & \\
    P3~\cite{BansalSEIP20} & \XSolid & & \XSolid\\
    P4~\cite{LiDSN14} & \XSolid & & \\    
    P5~\cite{LiRESS17} & \XSolid & \XSolid & \\
    P6~\cite{Hemmat16} & \XSolid & & \XSolid \\    
    P7~\cite{ZengTC21} & \XSolid & & \XSolid \\
    P8~\cite{ChenASE20} & \XSolid & & \\
    P9~\cite{JiangFSE20} & \XSolid & & \XSolid \\
    P10~\cite{ChenASE19} & \XSolid & & \XSolid \\
    P11~\cite{LinFSE18} & \XSolid & & \XSolid \\
\bottomrule
\end{tabular}
\end{table}

In order to understand whether the recent AIOps studies account for the three aforementioned key dimensions along the interpretation consistency, we conducted a literature survey of AIOps studies that derive interpretations from their AIOps models. To survey the literature, we searched Google Scholar with terms including  "AIOps", "Software Engineering", "Hard Drive Failures", and "Incident Prediction" to collect the initial set of studies. We then filtered these studies to keep the ones that were published in the last seven years (i.e., after 2014). In the end, eleven studies that we included in our survey were carefully examined by one of the authors. The results are shown in Table~\ref{tab:background_issues}.

According to our survey,  none of the surveyed papers explicitly mentioned that they control the randomness from the learners, data sampling, and hyperparameter tuning, where applicable. 2 out of 11 studies suggested interpreting low-performing models to derive interpretations. While 7 out of 11 studies did not consider or discuss the impact of periodic retraining strategies on the interpretation consistency. 

Our summary by no means intends to criticize the prior studies, but to raise awareness that there is currently a gap between recent AIOps studies and the commonly followed practices (e.g., controlling the randomness) in the machine learning community. It is important to take into account if the interpretations derived from AIOps models are internally consistent, externally consistent, or time consistent to avoid misleading decisions from being made by practitioners.
\subsection{Our criteria for assessing the consistency of AIOps interpretation}



Here we describe the motivation of assessing each dimension of AIOps model interpretation consistency in details and formulate the corresponding RQs.


\noindent\textbf{Internal Consistency.} Prior work has shown the impact of different random factors (e.g., random sampling of the same dataset, and different random seeds for fitting ML learner) in the reproducibility of predictive software engineering~\cite{Liemarxiv20} and deep learning~\cite{PhamASE20} research, with the focus on model performance reproducibility. As a result, the derived interpretations might not be reproducible (a.k.a internally consistent). Despite that, many of the AIOps studies did not take actions to ensure reproducibility. To investigate how the randomness from the three steps in the model training phase (highlighted in Figure~\ref{fig:background_pipeline}) impacts the internal consistency of interpretations derived from trained models, we formulate the following research question:

\vspace{2mm}
\noindent\fbox{
\parbox{0.98\linewidth} {\centering
\textit{\RQOne}
}
}
\vspace{2mm}

\noindent\textbf{External Consistency.} Many existing studies use interpretable models to compute feature importance ranks without taking into consideration the performance of these interpretable models. The assumption is that these interpretable models are able to capture the general trends in the data despite their lower performance. However, several studies already hint that such a practice could cause inconsistency in the computed insights~\cite{molnar2019,liptonQUEUE2018,rajbahadurTSE2020}. 
Hence, in the model evaluation phase (highlighted in Figure~\ref{fig:background_pipeline}), to assess the consistency of the interpretations derived from the evaluated models, we formulate the following research question: 

\vspace{2mm}
\noindent\fbox{
\parbox{0.98\linewidth} {\centering
\textit{\RQTwo}
}
}
\vspace{2mm}

\noindent\textbf{Time Consistency.} The workloads of large cloud computing platforms are highly dynamic and they tend to evolve significantly throughout its lifetime~\cite{LinFSE18,LiTOSEM20}. To keep up with the constantly evolving, temporal nature of the data, prior studies typically use several model update approaches to keep their AIOps models current and relevant. For instance, some studies periodically retrain their AIOps models on new data~\cite{LinFSE18,LiTOSEM20} or construct time-based ensemble approaches that aggregate local AIOps models trained on small time periods~\cite{streetKDD2001,wangKDD2003}. These models are typically interpreted to make operational decisions~\cite{LiTOSEM20,rudin2019stop}. However, the temporal nature of the data might make it hard for the different model update approaches to update the AIOps models to accurately generalize to the underlying trends while still yielding high performance on the most recent data. Hence, to avoid misleading operational decisions, it is important to prevent these updated AIOps models from losing sight of the historical trends in data and overfitting to the most recent time period. 

To examine which of the commonly used model updating approaches allows the interpretations of updated models to have the highest time consistency in the model evolution phase (highlighted in Figure~\ref{fig:background_pipeline}), we formulate the following research question:

\vspace{2mm}
\noindent\fbox{
\parbox{0.98\linewidth} {\centering
\textit{\RQThree}
}
}

\section{Case Study Setup}
\label{sec:case_study_setup}

In this section, we describe the setup of our case study in details.

\subsection{Studied AIOps Datasets} 
To understand the challenges of reliably interpreting AIOps models, we perform a case study on two large-scale public AIOps datasets that have been commonly used in prior work~\cite{LyuTOSEM21, BotezatuKDD16, SayedICDCS17, MahdisoltaniATC17, RosaIWQOS15, XuATC18}: the Google cluster trace dataset~\cite{reissGOOGLE2011} (hereinafter referred to as the Google dataset), and the Backblaze hard drive statistics dataset~\cite{backblazeDATA} (hereinafter referred to as the Backblaze dataset). 

\subsubsection{The Google dataset} 
The Google dataset contains the trace information of job runs on a large-scale cluster at Google. In this study, we use the second version of the dataset\footnote{\url{https://github.com/google/cluster-data/blob/master/ClusterData2011_2.md}} that was collected on May 2011, containing 29 days of trace information from a cluster of about 12.5K machines. The dataset includes information about the machines in the cluster, the jobs executed on the cluster, and the tasks under each job. In total, there are 670K jobs and 26M tasks in the dataset. There are four possible states of a job in its lifecycle: unsubmitted, pending, running, and dead. Throughout the lifecycle, a job triggers multiple events that are recorded in the dataset, including submit, schedule, evict, fail, kill, finish, lost, and update.

\subsubsection{The Backblaze dataset}
The Backblaze dataset contains daily snapshot of statistics of the hard drives in the Backblaze data center. The Backblaze dataset includes hard drive information (e.g., the model and capacity of the hard drive) and SMART (Self-Monitoring, Analysis and Reporting Technology) metrics of the hard drives. SMART is a hard drive monitoring system that monitors various indicators of drive reliability, to identify imminent hard drive failures. In this study, we use 36 months of Backblaze snapshot data collected from 2015 to 2017, containing 72M records.

\subsection{Experiment Context}
In this case study, we focus on two AIOps applications: cluster job failure prediction on the Google dataset, and hard drive failure prediction on the Backblaze dataset. The interpretation of cluster job failure prediction models can help practitioners identify factors that increase the likelihood of job failure, apply fail-safe mechanisms at the time of job submission, and investigate methods to reduce the occurrence of such risky factors. The interpretation of hard drive failure prediction models can help practitioners understand the early indicators of hard drive failures and improve monitoring infrastructure around the early indicators. Below we describe the general experiment context of our case study. We describe the RQ-specific experiment designs in the sections of each RQ. A replication package of our experiments and analysis are also provided~\footnote{\url{https://github.com/YingzheLyu/AIOpsInterpretation}. The replication package is set to be private during the review period, and will be made public once the paper is accepted. Please use the following GitHub account to view the repository in the meanwhile. Username: \texttt{reviewer-AIOpsInterpretation}, Password: \texttt{gUU!9sL9UDPT}}.

\subsubsection{Data preprocessing}~\label{sec:data_preprocessing}
Here we discuss the data preprocessing steps for two datasets separately.

\textbf{Google cluster job failure prediction:}
To predict job failures on the Google dataset, we need to first identify whether a job in the dataset finished successfully or not. Such information is not directly provided in the dataset. We use the events of jobs in the dataset as indications for such information. Specifically, we consider a job as failed if its last event in the dataset is a ``fail'' event. To avoid mislabelling jobs that have not finished running at the time of data collection, we exclude the jobs that started on the last day available in the dataset.

Similar to prior work~\cite{LyuTOSEM21, SayedICDCS17}, we extract a set of temporal and configuration metrics as candidate features for training the predictive models. Tantithamthavorn and Hassan~\cite{tantithamthavornICSE2018} stated that highly-correlated features should not be used to construct models for interpretation. Therefore, to remove the correlated and redundant features, we use the \texttt{varclus} function and the \texttt{redun} function in the \texttt{Hmisc} R package to identify and remove metrics that show high Spearman correlation and multicollinearity. Table~\ref{tab:features_google} provides a description of metrics used in our study for Google cluster job failure prediction.

We then divide the dataset into subsets with equal time intervals (i.e., periods), based on the date when the job is created. In total, we divide the dataset into 28 one-day periods.

\begin{table}
    \centering
    \caption{Description of metrics for Google cluster job failure prediction.}
    \label{tab:features_google}
    \resizebox{\textwidth}{!}{
    \begin{tabular}{p{12em}p{32em}}
        \hline
        Metric & Description\\
        \hline
        Scheduling class & A number that affects policies for resource access.\\
        Num Tasks & The number of tasks in a job.\\
        Priority & The priority of the job.\\
        Different machine & Whether a task must be scheduled to execute on a different machine than any other currently running task in the job.\\
        Requested CPU & Requested CPU resources.\\
        Requested Disk & Requested disk space resources.\\
        Mean CPU usage & Mean CPU usage over 5 minutes after job submission.\\
        Mean Memory usage & Mean memory usage over 5 minutes after job submission.\\
        Mean Disk Usage & Mean disk usage over 5 minutes after job submission.\\
        Sd Memory usage & Standard variation of the memory usage over 5 minutes after job submission.\\
        \hline
    \end{tabular}}
\end{table}

\textbf{Backblaze hard drive failure prediction:}
To predict hard drive failures on the Backblaze dataset, we extract the same set of metrics as prior work~\cite{LyuTOSEM21} as candidate features. The dataset is divided into subsets of one-month intervals (i.e., periods), to allow us label the hard drives that fail in the next period and associate them with their metrics in the current period. In total, we obtained 36 one-month periods, with the metrics of each hard drive in the period, and whether each hard drive fails in the next period. 

Similar to the Google dataset, we remove metrics that show high Spearman correlation and multicollinearity. Table~\ref{tab:features_disk} provides a description of metrics used in our study for Backblaze hard drive failure prediction. 

\begin{table}
    \centering
    \caption{Description of metrics for Backblaze disk failure prediction.}
    \label{tab:features_disk}
    \resizebox{\textwidth}{!}{
    \begin{threeparttable}
    \begin{tabular}{p{20em}p{25em}}
    \hline
    Metric & Description\\
    \hline
    Read Error Rate & Frequency of errors while reading raw data from a disk. \\
    Start/Stop Count~\tnote{1}& Number of spindle start/stop cycles.\\
    Reallocated Sectors Count & Quantity of remapped sectors.\\
    Seek Error Rate & Frequency of errors while positioning.\\
    Power-On Hours & Number of hours elapsed in the power-on state.\\
    Power Cycle Count & Number of power-on events.\\
    Reported Uncorrectable Errors & Number of reported uncorrectable errors, the definition is vendor-specific.\\
    Load Cycle Count & Number of cycles into landing zone position. \\
    HDA Temperature & Temperature of a hard disk assembly. \\
    Current Pending Sector Count & Number of unstable sectors (waiting for remapping). \\
    UltraDMC CRC Error Count & Number of CRC errors during UDMA mode. \\
    \hline
    \end{tabular}
    \begin{tablenotes}
    \item[1] For cumulative SMART attributes, we extract both their raw value from the last day and the difference during the training period as features, following the setup of \citet{LyuTOSEM21}.
    \end{tablenotes}
    \end{threeparttable}}
\end{table}

\subsubsection{Model training}~\label{sec:learners}
To ensure the result of our case study is generalizable, we include a variety of learners in our study that have been used in literature~\cite{BotezatuKDD16, SayedICDCS17, MahdisoltaniATC17, LyuTOSEM21}: Linear Discriminant Analysis (LDA), Quadratic Discriminant Analysis (QDA), Logistic Regression (LR), Classification And Regression Tree (CART), Gradient Boosting Decision Tree (GBDT), Random Forest (RF), and Multi-layer Perceptron Neural Network (NN). 

To mitigate the impact of different scales of metrics on model performance and interpretation, we perform data standardization on each metric in the training dataset by removing the mean of the metric and scaling the metric to its unit variance, using the \texttt{StandardScaler} function in the \texttt{scikit-learn} Python package.

We observe that the dataset is extremely imbalanced, with only 1\% failed jobs in the Google dataset and 0.1\% failed hard drives in the Backblaze dataset. To mitigate the impact of imbalanced dataset on model performance, we downsample the majority class (i.e., succeed jobs in the Google dataset and normal hard drives in the Backblaze dataset) in the training dataset to a success-to-fail ratio of 10:1 prior to training the model.

The detailed configuration of the model training process is described in sections of each RQ.

\subsubsection{Model evaluation}
\label{sec:setup-evaluation}
Prior work~\cite{tantithamthavornICSE2018} shows that one should use threshold-independent metrics such as Area Under the ROC Curve (AUC) in lieu of threshold-dependent metrics such as Precision, Recall or F-measure to evaluate model performance. Therefore in this case study, we use AUC to evaluate the performance of each trained model.

Because of the temporal nature of AIOps datasets, traditional evaluation methods such as cross validation may result in data leakage and therefore lead to inaccurate performance evaluation~\cite{LyuTOSEM21}. In this case study, we evaluate a model that is trained on a specific period of dataset using data in the next period as the testing dataset. 

To prepare the testing dataset, we apply the same data scaler that was fitted on the training dataset on the testing dataset. It is worth noting that although we rebalance the training dataset by downsampling the majority class, we do not perform such downsampling on testing dataset.

\subsubsection{Model interpretation}
There are two types of model interpretation: model-level interpretation (i.e., identifying the features that have the biggest impact on a model's predictions), and instance-level interpretation (i.e., identifying the reason that a model predicts a specific input to be a specific outcome). In this case study, we focus on model-level interpretation because prior work on AIOps mostly focus on model-level interpretation~\cite{LiTOSEM20, BansalSEIP20, LiDSN14, LiRESS17, ChenASE19, JiangFSE20}.

In particular, we use permutation feature importance, a model-agnostic approach to derive the importance of features in a machine learning model in our study. Permutation feature importance evaluates the importance of a feature by randomly shuffling the value of the feature on the testing dataset, and measure the drop of model performance due to such shuffling. We use permutation feature importance in our case study because: (1) some of our studied learners are not intrinsically interpretable; (2) permutation feature importance offers a consistent way to compare across all the models; and (3) many existing studies use permutation feature importance on the interpretations of machine learning models~\cite{rajbahadurTSE2020,tantithamthavorn2018TSE2}. To control the impact of randomness in permutation feature importance calculation on our case study results, we fix the random seed of the permutation feature importance calculation.

In the rest of the paper, we will use the term interpretation and feature importance ranks interchangeably.

\subsubsection{Similarity measurement of model interpretation}
To compare the similarity of interpretations among two or more models, we use the following measurements:

\begin{itemize}
    \item \textbf{Kendall's Tau}: a non-parametric measure of the similarity between two rankings. Kendall’s Tau ranges from 0 (no agreement) to 1 (complete agreement). We use the same interpretation schema from prior work~\cite{rajbahadurTSE2020} to interpret Kendall’s Tau in our study: 
    \[
    \text{Kendall's Tau Agreement} = 
    \left\{ \begin{array}{ll}
      \text{Weak}      &  \text{if } 0 \leq Tau\leq 0.3. \\
      \text{Moderate}  &  \text{if } 0.3 < Tau \leq 0.6. \\
      \text{Strong}    &  \text{if } 0.6 < Tau \leq 1.
   \end{array} \right.
    \]

    \item \textbf{Kendall's W}: a non-parametric measure of the similarity among multiple rankings. Similar to Kendall’s Tau, Kendall's W also ranges from 0 (no agreement) to 1 (complete agreement). We use the same interpretation schema that we use for Kendall’s Tau. 
    
    \item \textbf{Top K Overlap Score}: after filtering out features with a negligible importance score (i.e., permutation feature importance score lower than 0.0001), we use the same definition of the Top K Overlap Score as prior work~\cite{rajbahadurTSE2020}:
    \[\text{Top K Overlap Score} = \frac{\cap_{i\geq2}^n\text{ Most important K features for model }i}{\cup_{i\geq2}^n\text{ Most important K features for model }i}\] where $n$ is the number of models for comparison. 
\end{itemize}

\section{\RQOne}
\label{sec:rq1}

In this section, we evaluate the internal consistency of AIOps model interpretation.


\subsection{Approach.}
Figure~\ref{fig:background_pipeline} shows three components involved in the model training phase of AIOps pipeline: data sampling, hyperparameter tuning, and ML learner. Prior work~\cite{Liemarxiv20, PhamASE20} shows that randomness in the model training phase leads to inconsistent model performance. In this RQ, we hypothesize that randomness in the model training phase also leads to inconsistent model interpretations. In particular, the fitting algorithm of the ML learner may have inherent randomness; the randomized hyperparameter tuning contains randomness; and the random sampling of training dataset also introduces randomness to the model training phase~\cite{PhamASE20}.

To validate our hypothesis, we conduct a set of controlled experiments in this RQ: for each of the potential sources of randomness, we control for the other sources of randomness and train the 7 studied learners on the two datasets. More specifically:

\begin{enumerate}
    \item To evaluate the impact of the inherent randomness from the learner on the internal consistency of AIOps model interpretations, we control for the hyperparameters and the training dataset, i.e., we fix the hyperparameters of each learner to its default hyperparameters, and train a model for each studied learner using all data in a given time period and dataset, without controlling for the random seed of the learner's fitting algorithm. The process is repeated 10 times, and the interpretation similarity measurements among the 10 iterations are calculated for each learner on each dataset and its each time period.
    \item To evaluate the impact of the randomness from randomized hyperparameter tuning on the internal consistency of AIOps model interpretations, we control for the inherent randomness from the learner and the training dataset, i.e., for each studied learner, we train a model with randomized search on optimal hyperparameters, using all data in a given time period and dataset, with a fixed random seed for the learner's fitting algorithm. The process is repeated 10 times for each dataset and its each time period, and the interpretation similarity measurements among the 10 iterations are calculated for each learner on each dataset and its each time period.
    \item To evaluate the impact of the randomness from data sampling, we control for the inherent randomness from the learner and the hyperparameters, i.e., for each studied learner, we fix the hyperparameters of the learner to its default hyperparameters, and the random seed for the learner's fitting algorithm, and train a model using a bootstrapped sample in a given time period and dataset. Unlike in Experiment 1, 3, and 4 where we use the same training data (i.e., all data in a given time period) for all iterations to control for training data randomness, here through bootstrapping a different sample of training data in each iteration to intentionally introduce data randomness. The process is repeated 10 times for each dataset and its each time period, and the interpretation similarity measurements among the 10 iterations are calculated for each learner on each dataset and its each time period.
    \item Finally, to evaluate if we can derive an internally consistent interpretation of a AIOps model when controlling for all the above three factors, we conduct a fully controlled experiment: we fix the hyperparameters of each learner to its default hyperparameters, and train a model for each studied learner using all data in a given time period and dataset, with a fixed random seed for the learner's fitting algorithm. The process is repeated 10 times, and the interpretation similarity measurements among the 10 iterations are calculated for each learner on each dataset and its each time period.
\end{enumerate}

Table~\ref{tab:rq1_setup} shows an overview of our controlled experiment setup.

\begin{table}
\centering
\caption{Controlled experiment setup for RQ1.}
\label{tab:rq1_setup}
\resizebox{0.65\textwidth}{!}{
\begin{threeparttable}
\begin{tabular}{cccc}
\toprule
\multirow{2.5}{*}{Experiment}    & \multicolumn{3}{c}{Potential source of randomness}  \\
\cmidrule(l){2-4}
 & ML learner     & Hyperparameter tuning & Data sampling  \\
\midrule
1          & \XSolid        & \Checkmark            & \Checkmark     \\
2          & \Checkmark     & \XSolid               & \Checkmark     \\
3          & \Checkmark     & \Checkmark            & \XSolid \\
4          & \Checkmark     & \Checkmark            & \Checkmark     \\
\bottomrule
\end{tabular}
\begin{tablenotes}
\small
\item \Checkmark~~-~~Controlled;  \XSolid~~-~~Not Controlled.
\end{tablenotes}
\end{threeparttable}}
\end{table}

\subsection{Results.}

\begin{figure}
    \centering
    \includegraphics[width=\textwidth]{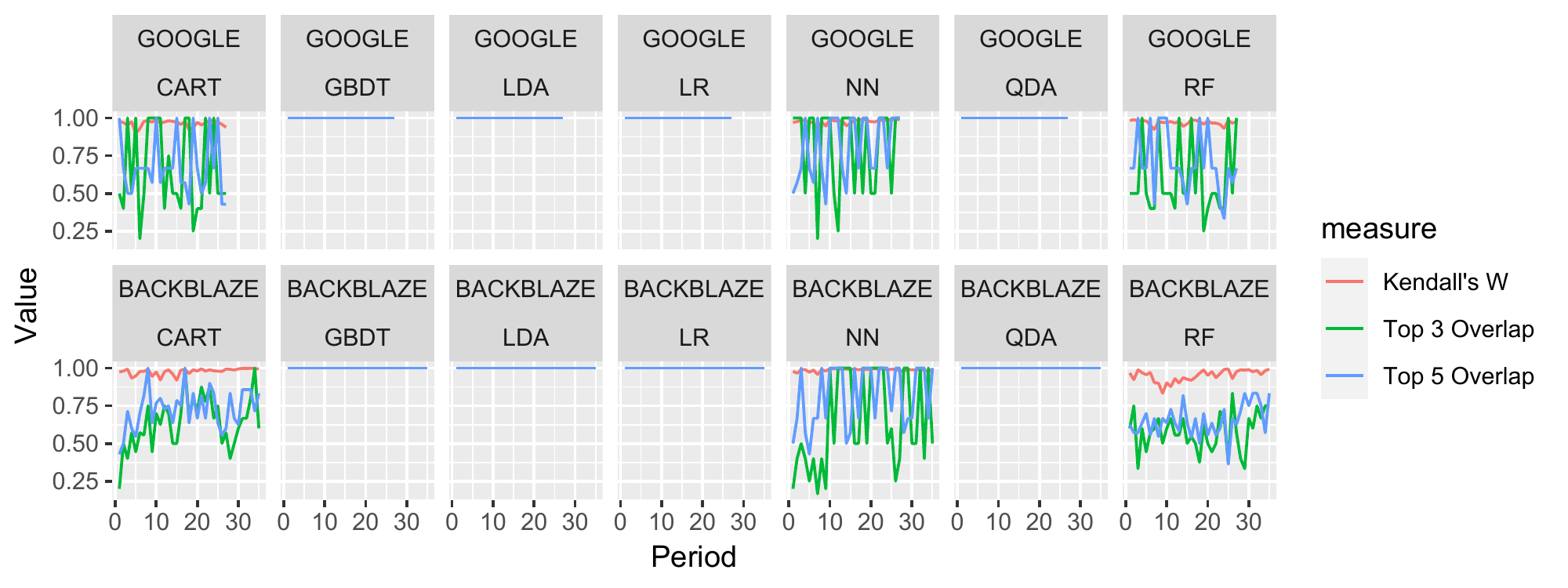}
    \caption{Consistency of the AIOps model interpretations across iterations under the impact of inherent randomness from learners.}
    \label{fig:rq1.1}
\end{figure}

\textbf{Inherent randomness from learners introduces inconsistencies to the derived interpretations of an AIOps model.} Figure~\ref{fig:rq1.1} shows the interpretation consistency among iterations when controlling for the hyperparameters and the training dataset. As shown in Figure~\ref{fig:rq1.1}, learners such as NN, RF and CART show inconsistent interpretation among iterations, while LDA, QDA, LR and GBDT do not show inconsistent interpretation among iterations. It is worth noting that the degree of inherent randomness of a learner's fitting algorithm depends on the specific implementation of such algorithm. While scikit-learn’s implementation of LDA and QDA do not involve random states~\footnote{\url{https://github.com/scikit-learn/scikit-learn/blob/15a949460/sklearn/discriminant_analysis.py}}. Scikit-learn provides multiple solvers for LR, among which some involve randomness (\texttt{sag}, \texttt{saga} or \texttt{liblinear}) while others don’t~\footnote{\url{https://github.com/scikit-learn/scikit-learn/blob/15a949460/sklearn/linear_model/_logistic.py}}. Similarly, xgboost’s implementation of GBDT supports multiple boosters and only shows non-deterministic behavior when a gblinear booster (i.e., Hogwild algorithm) is used~\footnote{\url{https://xgboost.readthedocs.io/en/latest/python/python_api.html}}. In our study, we used the default solver for LR (i.e., \texttt{lbfgs}) and the default booster for GBDT (i.e., \texttt{gbtree}), both of which do not involve randomness. Hence, the apparent stability in Figure~\ref{fig:rq1.1} may be dependent on the implementation of the learners that we used and may not be generalizable to other implementations.

\begin{figure}
    \centering
    \includegraphics[width=\textwidth]{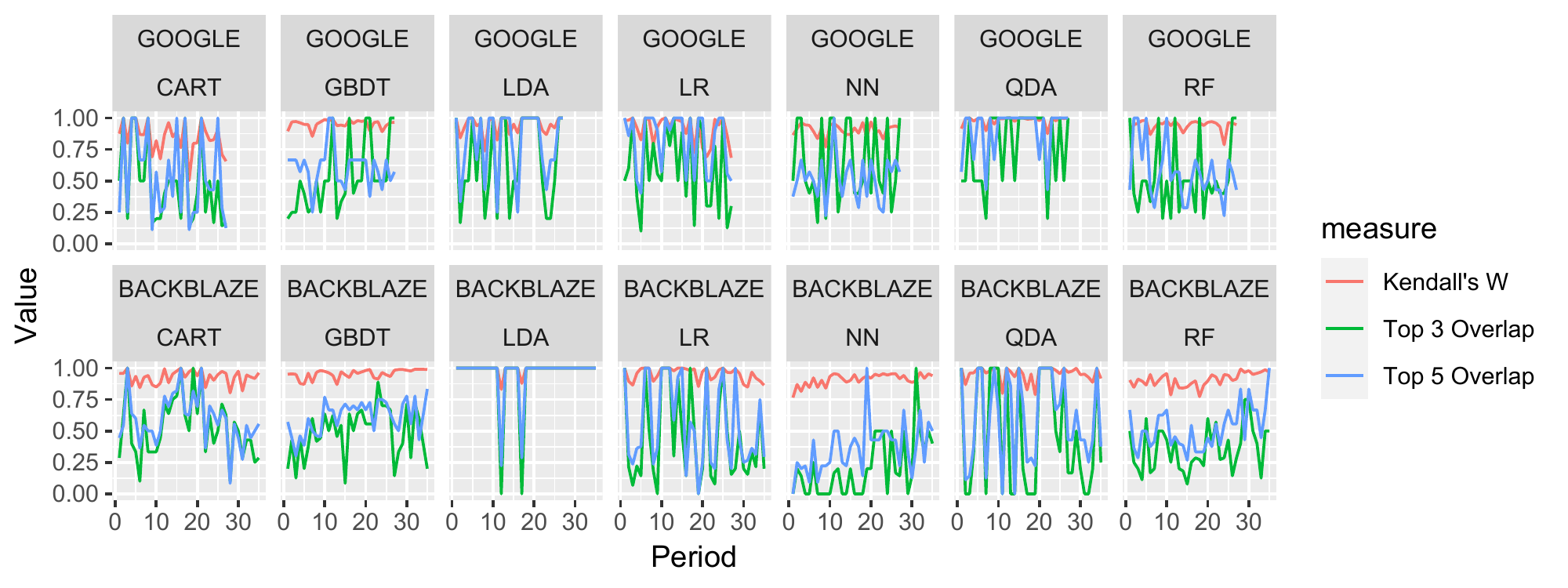}
    \caption{Consistency of the AIOps model interpretations across iterations under the impact of randomized hyperparameter searching.}
    \label{fig:rq1.2}
\end{figure}

\textbf{Randomized hyperparameter searching introduces inconsistencies to the interpretation of AIOps models.} Figure~\ref{fig:rq1.2} shows the interpretation consistency among iterations when controlling for the inherent randomness from the learner and the training dataset. As shown in Figure~\ref{fig:rq1.2}, randomized hyperparameter searching introduces inconsistent interpretation among iterations across all studied learners.

\begin{figure}
    \centering
    \includegraphics[width=\textwidth]{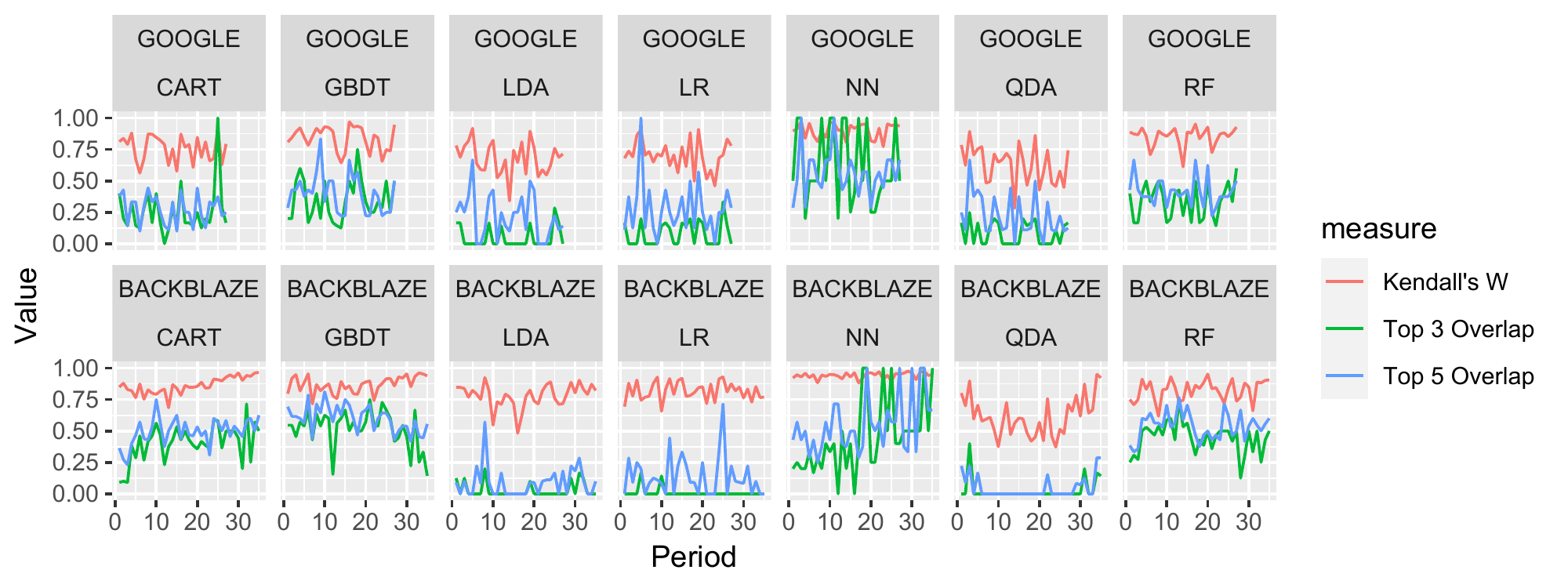}
    \caption{Consistency of the AIOps model interpretations across iterations under the impact of sampling randomness.}
    \label{fig:rq1.3}
\end{figure}

\textbf{Sampling randomness introduces inconsistencies to the interpretation of AIOps models.}  Figure~\ref{fig:rq1.3} shows the interpretation consistency among iterations when controlling for the inherent randomness from the learner and the hyperparameters. As shown in Figure~\ref{fig:rq1.3}, sampling (with bootstrap) introduces inconsistent interpretation among iterations across all studied learners.

\begin{figure}
    \centering
    \includegraphics[width=\textwidth]{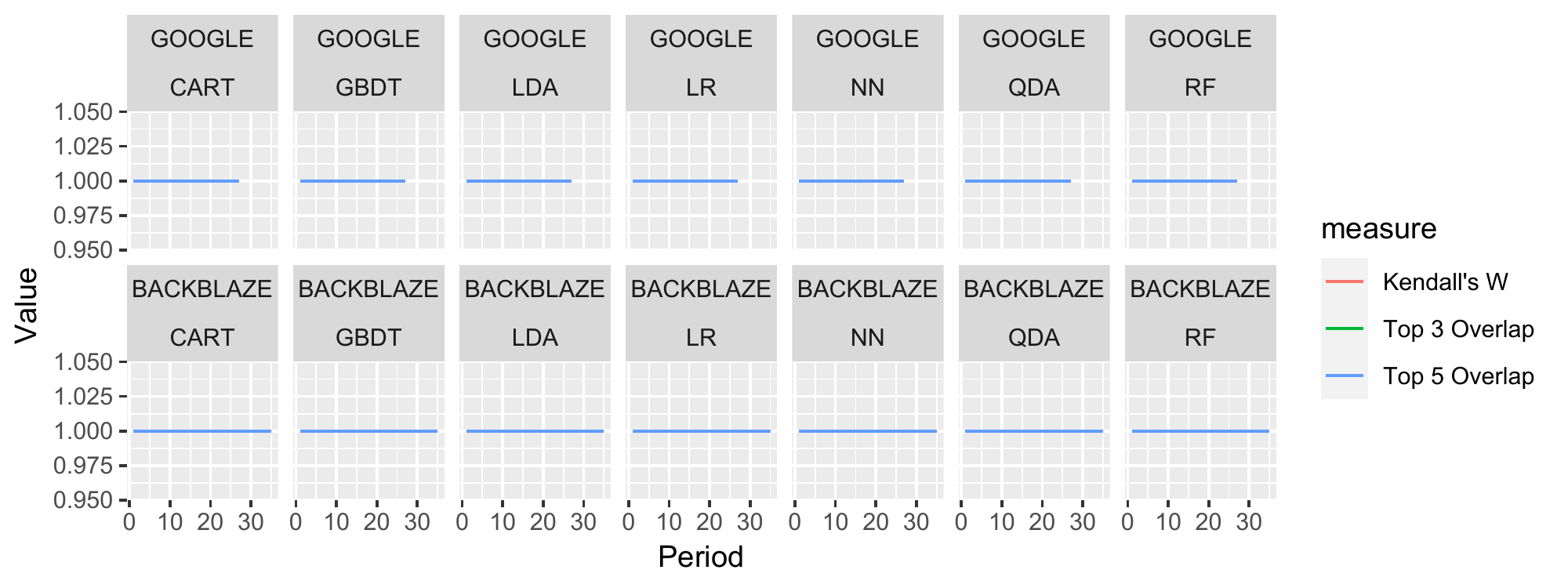}
    \caption{Consistency of the AIOps model interpretations across iterations when controlling all three sources of randomness.}
    \label{fig:rq1.4}
\end{figure}

\textbf{When controlling all three sources of randomness, AIOps model interpretations are internally consistent.} Figure~\ref{fig:rq1.4} shows the interpretation consistency among iterations when controlling for all three sources of randomness. As shown in Figure~\ref{fig:rq1.4}, all learners in every period of both datasets yield identical interpretations across iterations. The result indicates that when the randomnesses of learner, data sampling, and hyperparameter tuning are all controlled, AIOps model interpretation becomes internally consistent.

\begin{figure}
    \centering
    \includegraphics[width=\textwidth]{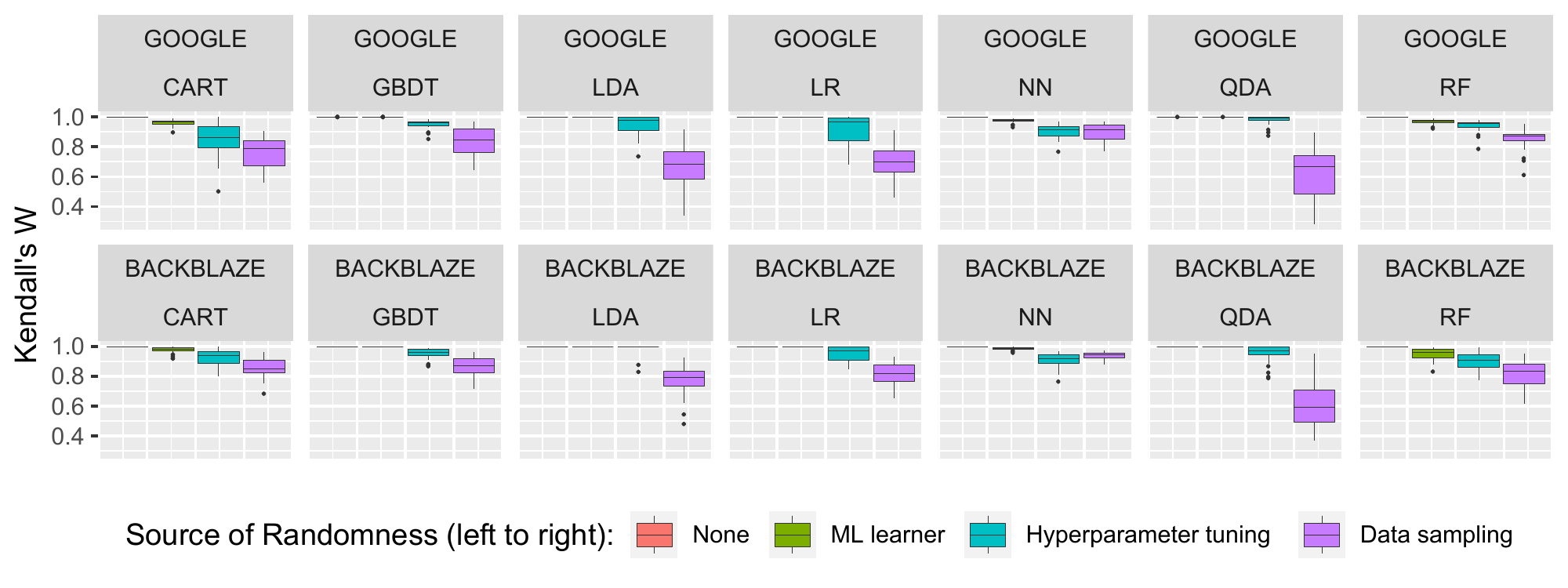}
    \caption{The interpretation consistency (measured by Kendall's W) of models when different sources of randomness are not controlled. Each subplot has four boxplots, from left to right represents the control group (all sources of randomness controlled), randomness from ML learner, hyperparameter tuning, and data sampling. See Table~\ref{tab:rq1_setup} for detailed setup.}
    \label{fig:discussion_randomness}
\end{figure}

\textbf{The randomness from data sampling introduces the largest scale of inconsistency to AIOps model interpretation.} Figure~\ref{fig:discussion_randomness} provides an aggregated view of the distribution of Kendall's W when different sources of randomness are not controlled, in comparison to a controlled group with all sources of randomness controlled. It can be observed from Figure~\ref{fig:discussion_randomness} that generally when randomness from data sampling is introduced, the interpretation of AIOps models show the lowest consistency; followed by hyperparameter tuning which shows the second lowest consistency among model interpretations. The ML learner's inherent randomness appears to introduce the least inconsistency to AIOps model interpretation. Such observation is consistent across most learners and datasets, except for Multi-layer Perceptron Neural Network (NN) - in the case of NN, hyperparameter tuning introduces comparable interpretation inconsistency to data sampling. A possible explanation is that for NN, the architecture of the network (e.g., the size of hidden layers) are tuned as a group of hyperparameters, which may result in significant difference in the architecture of the trained models. Because of the wide adoption of Neural-Network-based learners in AIOps research (e.g., Deep Neural Network, DNN), future work is needed to further explore the impact of hyperparameter tuning randomness on DNN model interpretation.

\vspace{0.5cm}
\begin{Summary}{Summary of RQ1}{}
Inherent randomness from learners, randomized hyperparameter searching, and sampling randomness all result in AIOps models yielding different interpretations across different repeated runs. Sampling randomness introduces the largest scale of inconsistency to AIOps model interpretation. By controlling the randomness from the learner, hyperparameter tuning, and data sampling, we are able to derive internally consistent interpretations for AIOps models. 
\end{Summary}
\section{\RQTwo}
\label{sec:rq2}

In this section, we evaluate the external consistency of AIOps model interpretation.



\subsection{Approach.}


Our approach to evaluate the external consistency of AIOps model interpretation consists of three steps: (1) model generation; (2) model clustering; and (3) interpretation comparison.

\textbf{Step 1: Model generation.} In Section~\ref{sec:rq1}, we find that the randomness existing in hyperparameter tuning, data sampling, and ML learners themselves would lead to a large variety of models being generated.
To preserve the randomness and generate models in different performance scales, we train models on bootstrapped dataset samples, with randomized hyperparameter searching. For each of the learners at each period, we repeat the model training process for 10 iterations. During each iteration, a bootstrapped dataset sample is randomly generated without setting explicit random seeds.
We apply this process to both studied datasets. In the end, for each dataset and each period, we generated 70 models. For example, the Google dataset contains data of 28 periods, hence we generated $27 \times 70 = 1,890$ models. We did not use the data from the last period for training as there would be no available data for testing the performance of models.

\textbf{Step 2: Model clustering.} In this step, we cluster the models trained with the data in same periods based on the model performance (i.e., the AUC metric). 
We use a widely-adopted one-dimensional clustering technique called Jenks natural breaks optimization~\cite{Jenks1967TheDM}. To decide the optimal number of clusters, we first conducted the elbow method. The elbow method~\cite{ThorndikePsy53} visualizes the relationship between the number of clusters and the evaluation metric. It is a heuristic commonly used in cluster analysis to decide the optimal number of clusters. The elbow method considers the number of clusters to be optimal, if increasing the number of clusters has little impact on the evaluation metrics. In this study, we use Within Sum of Square (WSS) as the evaluation metric. WSS is to measure the variability of observations in each cluster and is calculated based on the clustering technique (i.e., Jenks natural breaks optimization in our case). As the number of clusters increases, the WSS will decrease as the variation in each cluster decreases. The optimal cluster number is the cutoff point where increasing cluster number has little impact on WSS.

\begin{figure}
    \centering
    \includegraphics[width=\textwidth]{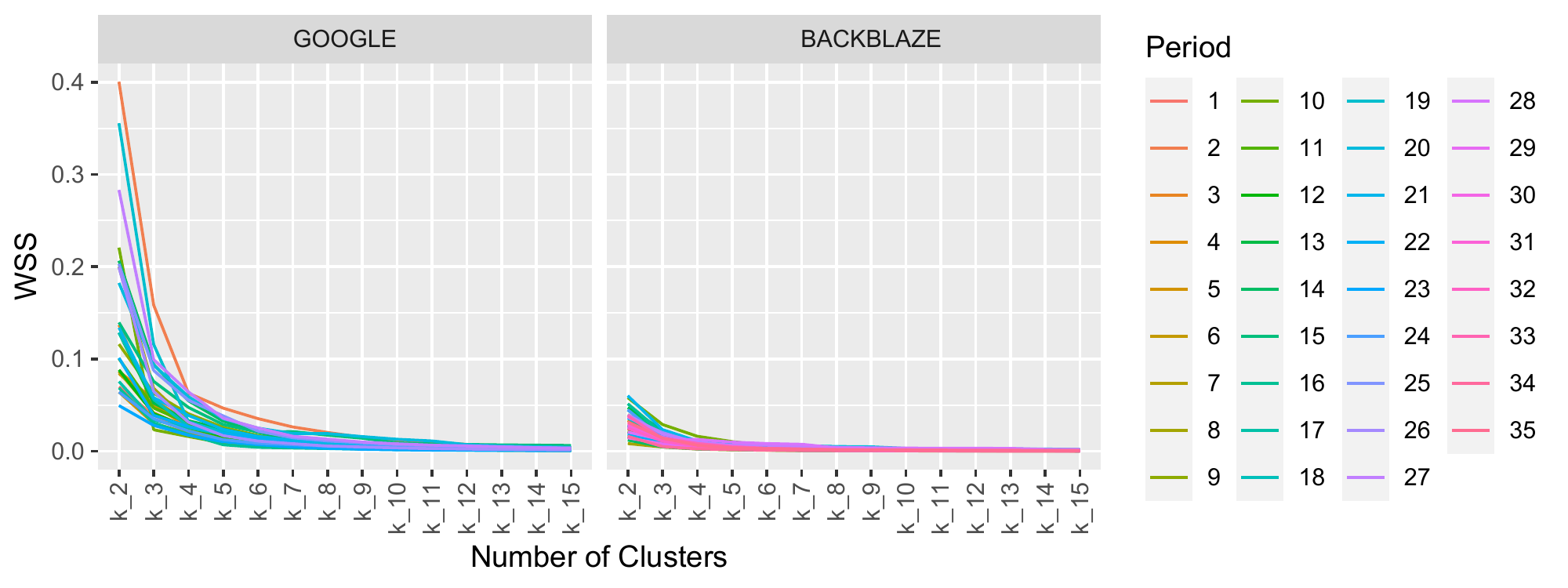}
    \caption{Elbow graph of performance clustering for deciding the optimal number of clusters.}
    \label{fig:rq2_elbow}
\end{figure}

Figure~\ref{fig:rq2_elbow} shows the relations between WSS and the number of clusters for both Google and Backblaze datasets. The x-axis shows the parameters we use in the Jenk natural breaks optimization. For example, $k\_2$ represents the final number of cluster is one. From the visualization, we decide that the optimal number of clusters for the Google dataset is 4 (i.e., $k\_5$ is the elbow point) and the optimal number of clusters for the Backblaze dataset is 2 (i.e., $k\_3$ is the elbow point).

We then apply Jenks natural breaks optimization to the 70 models in each period of each dataset, with the chosen optimal number of clusters, to group the models into different clusters based on their performance. We noticed that Jenks natural breaks cannot divide the models in Backblaze period 12 into two clusters with at least 2 models in each cluster. We therefore removed Backblaze period 12 from our analysis in the rest of this RQ.

\textbf{Step 3: Interpretation comparison.} In this step,  we calculate the interpretation similarity measurements within each of the clusters to check if the interpretations of models in the same performance group are externally consistent. As described in Section~\ref{sec:case_study_setup}, we use three types of similarity metrics to measure the consistency of interpretations: the Kendall's W, Top 3 overlap score, and Top 5 overlap score.

To statistically compare the interpretation consistency in different performance groups, we perform the Wilcoxon Rank Sum (WRS) test. The Wilcoxon Rank Sum test is an unpaired, non-parametric test commonly used in literature~\cite{ChenEMSE2019, LinEMSE2018, LinEMSE2019}, of which the null hypothesis is that for randomly selected values X and Y from two distributions, the probability of X being greater than Y is equal to the probability of Y being greater than X. In this RQ, to confirm if the interpretation consistency of group X is statistically significantly higher than group Y, we use a one-sided alternative hypothesis of X being shifted to the right of Y. Therefore, if the p-value of the Wilcoxon Rank Sum test is less than 0.05, we conclude that group X has more consistent interpretations within the group than group Y. 

To mitigate the threat of false positives during multiple comparisons, we use Bonferroni correction~\cite{Miller2012} to correct the p-value before concluding a Wilcoxon Rank Sum test where applicable.

The Wilcoxon Rank Sum test only shows whether two distributions are different, but not the magnitude of the difference. To assess the magnitude of the differences, we also calculate the effect size using Cliff’s Delta $d$. We use the following schema for interpreting $d$, which is widely used in prior work~\cite{ChenEMSE2019, LinEMSE2018, LinEMSE2019}:
\[
    \text{Cliff's Delta }d = 
    \left\{ \begin{array}{ll}
      \text{Negligible (N)} &  \text{if }0 \leq |d| \leq 0.147. \\
      \text{Small (S)}      &  \text{if }0.147 < |d| \leq 0.33. \\
      \text{Medium (M)}     &  \text{if }0.33 < |d| \leq 0.474. \\
      \text{Large (L)}      &  \text{if }0.474 < |d| \leq 1.
   \end{array} \right.
\]

\subsection{Results.}

\begin{figure}
    \centering
    \includegraphics[width=\textwidth]{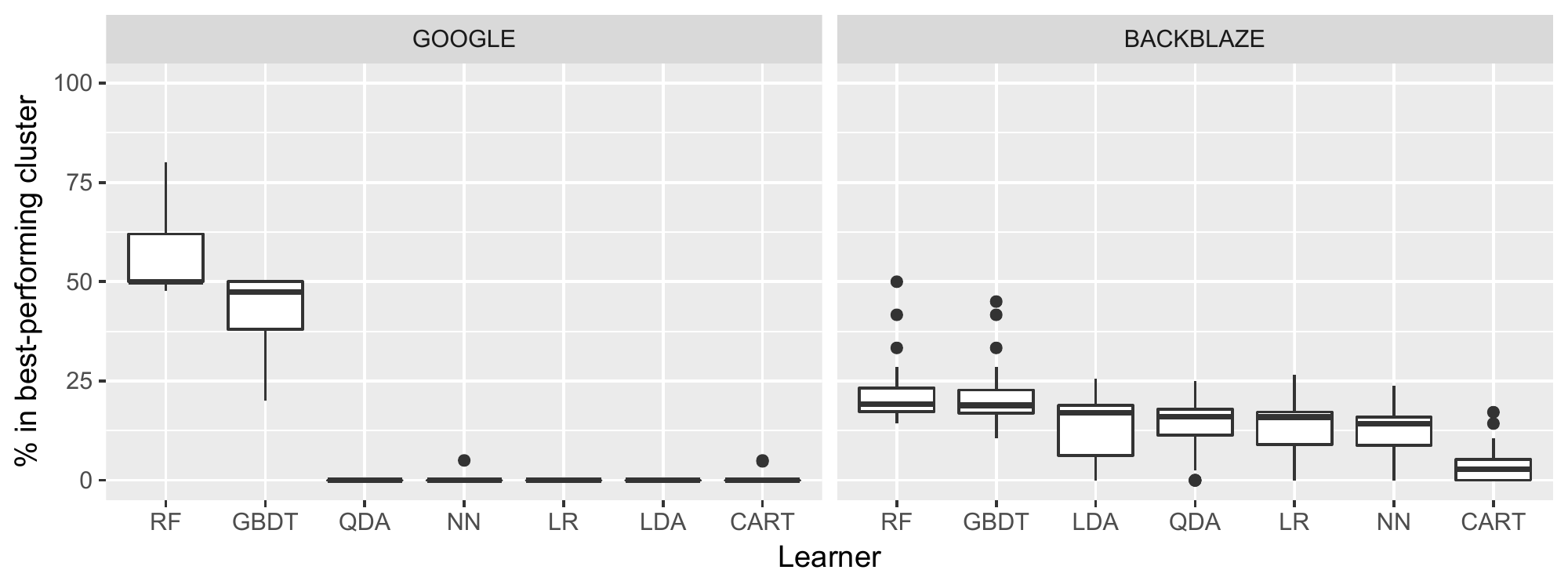}
    \caption{Percentage of models trained with each learner in the best-performing clusters.}
    \label{fig:discussion_top_learner}
\end{figure}

\noindent\textbf{RF and GBDT are the top two best-performing learners for the studied AIOps context.} Figure~\ref{fig:discussion_top_learner} shows the percentage of models trained with each learner in the best-performing clusters. For the Google dataset, Figure~\ref{fig:discussion_top_learner} shows that RF and GBDT perform much better than other learners. The median percentage of RF and GBDT are around 50\% and the percentage of other five learners are all below 1\%. 
Similarly, the results on the Backblaze dataset show that RF and GBDT are still the top two learners among all seven learners. Different from the results on the Google dataset, the performance of models trained using LDA, QDA, LR, and NN are comparable to those trained using RF and GBDT. In general, none of the other five learners generate more than 25\% of models in the best-performing cluster. 

The findings suggest that we should consider using RF and GBDT as the preferred learners in the context of studied AIOps applications and datasets, as they consistently perform the best among all studied learners. In addition, the performance of learners may vary due to different characteristics of different datasets. Future work should further explore the correlation between the AIOps dataset characteristics and performance of different learners.

\begin{figure}
    \centering
    \includegraphics[width=\textwidth]{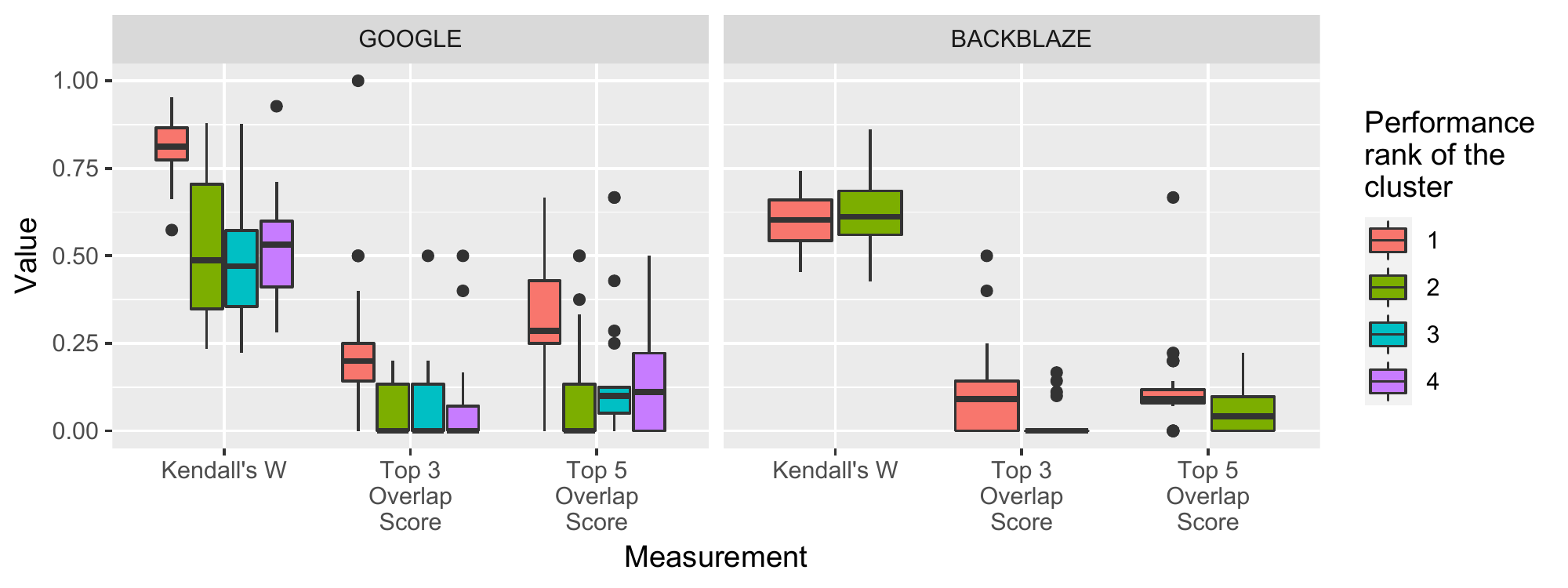}
    \caption{Distribution of similarity measurements for clusters with different performance ranks.}
    \label{fig:rq2_similarity_measurement}
\end{figure}

\noindent\textbf{Models in the highest-performed clusters tend to have more consistent interpretation results compared to lower-performed clusters.}
Figure~\ref{fig:rq2_similarity_measurement} shows the three types of similarity measurements for clusters in different performance ranks.

For the Google dataset, the top 1 ranked cluster has higher values in all the three measurements compared to other clusters. For example, the median of Kendall's W in top 1 ranked cluster is 0.81, while the median of Kendall's W in other clusters is only between 0.47 and 0.53. Table~\ref{tab:rq2_similarity} presents the detailed results of the comparison. For all three types of measurements, the values of the cluster in rank 1 are statistically higher than the clusters in rank 2, 3, and 4 (corrected p-value \textless 0.05). Furthermore, the results show that all the effect sizes are large. We also observe that the highest-performed cluster in the Google dataset have much more consistent interpretations.

For the Backblaze dataset, there are only two performance clusters: rank 1 and rank 2. We observe that, for the measurements of top 5 overlap score and top 3 overlap score, the differences between rank 1 cluster and rank 2 cluster are significant different (corrected p-value \textless 0.05). The magnitude of differences are either small or medium. On the other hand, for the measurement of Kendall's W, we find that the difference is insignificant. We further investigate the reason for insignificant results of Kendall's W. We notice that the AUCs of the models trained on Backblaze dataset are mostly above 0.75. Based on such observations, we hypothesize that the insignificant difference may be due to the fact that all the trained models have relatively high performance. To verify our hypothesis, we further explore the relationship between performance (i.e., AUC) and within-cluster interpretation similarity measurements (e.g., Kendall's W).

\begin{table}
\centering
\caption{Comparing the similarity measurement among clusters in different performance ranks.}
\label{tab:rq2_similarity}
\resizebox{0.9\textwidth}{!}{
\begin{threeparttable}
\begin{tabular}{llrrrr}
\toprule
\textbf{Measurement} & \textbf{Dataset} & \textbf{Rank X} & \textbf{Rank Y} & \textbf{\begin{tabular}[c]{@{}r@{}}Corrected p-value\\for WRS$^+$\end{tabular}} & \textbf{Cliff's $d^*$} \\
\midrule
\multirow{4}{*}{\textbf{Kendall's W}} & \multirow{3}{*}{Google} & 1 & 2 & < 0.01 & 0.71 (L) \\
 &  & 1 & 3 & < 0.01 & 0.84 (L) \\
 &  & 1 & 4 & < 0.01 & 0.89 (L) \\
 \cline{2-6}
 & Backblaze & 1 & 2 & 0.91 & - (N/A) \\
 \hline
 
\multirow{4}{*}{\textbf{Top 5 Overlap Score}} & \multirow{3}{*}{Google} & 1 & 2 & < 0.01 & 0.73 (L) \\
 &  & 1 & 3 & < 0.01 & 0.69 (L) \\
 &  & 1 & 4 & < 0.01 & 0.74 (L) \\
  \cline{2-6}
 & Backblaze & 1 & 2 & 0.01 & 0.32 (S) \\
 \hline
\multirow{4}{*}{\textbf{Top 3 Overlap Score}}  & \multirow{3}{*}{Google} & 1 & 2 & < 0.01 & 0.59 (L) \\
 &  & 1 & 3 & < 0.01 & 0.55 (L) \\
 &  & 1 & 4 & < 0.01 & 0.60 (L) \\
  \cline{2-6}
 & Backblaze & 1 & 2 & < 0.01 & 0.45 (M) \\
 \bottomrule
\end{tabular}
\begin{tablenotes}
\small
\item $^+$ $p < 0.05$ - Significant; $p \geq 0.05$ - Insignificant.
\item $^*$ N/A - Not Applicable; N - Negligible; S - Small; M - Medium; L- Large.
\end{tablenotes}
\end{threeparttable}}
\end{table}

\begin{figure}
    \centering
    \settowidth{\imagewidth}{\includegraphics{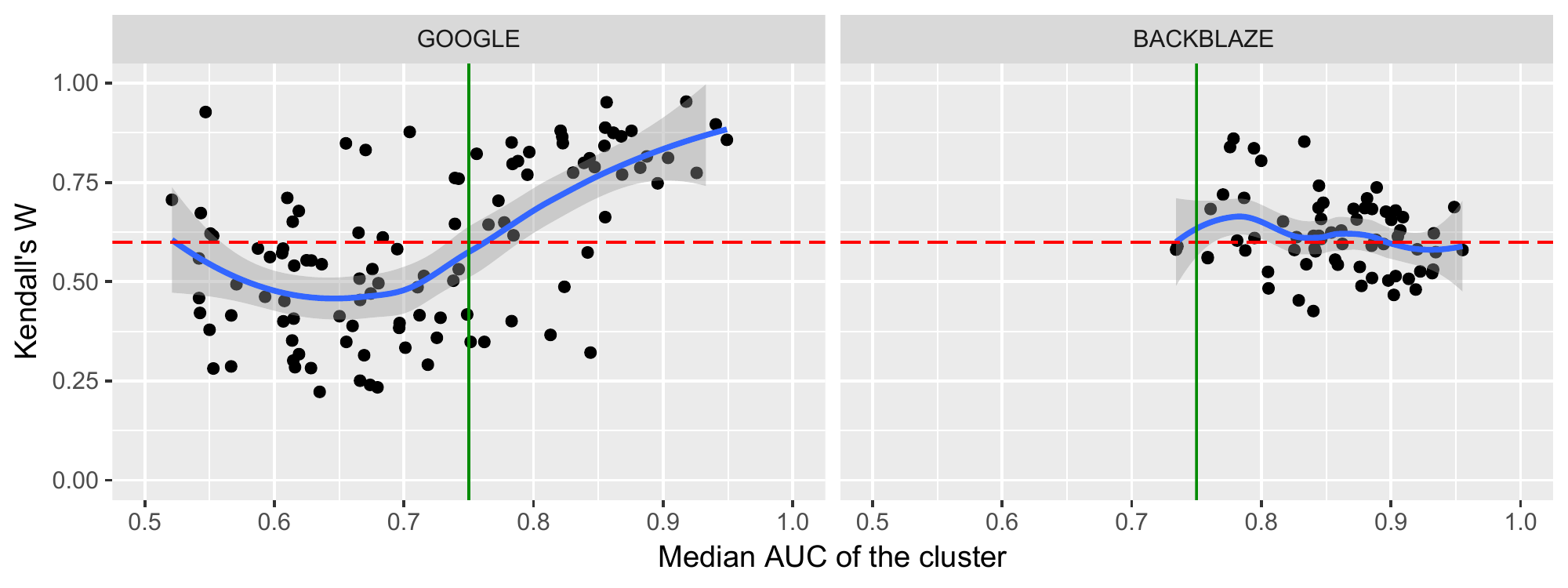}}
    \begin{subfigure}{.55\textwidth}
      \centering
      \includegraphics[trim=0 0 0.45\imagewidth{} 0, clip, width=\textwidth]{pics/RQ2_converge.pdf}
      \caption{}
      \label{fig:rq2_converge_google}
    \end{subfigure}\hspace{0em}%
    \begin{subfigure}{.45\textwidth}
      \centering
      \includegraphics[trim=0.55\imagewidth{} 0 0 0, clip, width=\textwidth]{pics/RQ2_converge.pdf}
      \caption{}
      \label{fig:rq2_converge_backblaze}
    \end{subfigure}
    \caption{Relationship between a cluster’s median AUC value and its within-cluster interpretation similarity, measured by Kendall’s W. The red dashed line marks a Kendall's W of 0.6 (a strong agreement), while the green solid line marks an AUC of 0.75. The blue curve is fitted using a Local Polynomial Regression and the grey area represents a 95\% confidence interval of the fit.}
    \label{fig:rq2_converge}
\end{figure}

\noindent\textbf{Models that have an AUC of at least 0.75 tend to have high consistency among their interpretation.} Figure~\ref{fig:rq2_converge} shows the relationship between a cluster's median AUC value and its within-cluster interpretation consistency represented by Kendall's W value. The blue line is fitted using a Local Polynomial Regression and the grey area represents a 95\% confidence interval. The red line indicate the threshold of Kendall's W (0.6) where the consistency is considered to be high.
Note that for the Backblaze dataset, all clusters have a median AUC above 0.70, with the majority of them above 0.75, while for Google dataset, the median AUCs span across 0.5 to 0.95. 
For the figure of the Google dataset in Figure~\ref{fig:rq2_converge_google}, we observe that the within-cluster interpretation similarity begins to increase after AUC is over 0.7, and exceeds a Kendall’s W of 0.6 when AUC is larger than 0.75. 

From Figure~\ref{fig:rq2_converge_backblaze}, we observe that most of the clusters' median AUC values are higher than 0.75 and Kendall's W is consistently above the high agreement threshold. While it is unclear whether the lower AUC values can still result in high Kendalls'W (i.e., > 0.6), from the observable trends in both datasets, we draw a conservative conclusion that models that have an AUC of at least 0.75 tend to have high consistency among their interpretations, and their interpretations are more reliable than models with AUCs less than 0.75.

\vspace{0.5cm}
\begin{Summary}{Summary of RQ2}{}
RF and GBDT are the top two best-performing learners for the studied AIOps context. Models in the best performing cluster tend to have more externally consistent interpretations compared to other groups. When the AUC scores are greater than 0.75, the interpretations of AIOps models are externally consistent. 

\end{Summary}
\section{\RQThree}
\label{sec:rq3}

In this section, we evaluate the time consistency of AIOps model interpretation.

\subsection{Approach.}

\begin{algorithm}[t]
\caption{Streaming Ensemble Algorithm (SEA)}
\small
\begin{algorithmic}[1]
\STATE $no\_{of}\_{learners}\_{in}\_{ensemble} \longleftarrow 0$
\STATE $E \longleftarrow \emptyset$ $E$ is the ensemble 
\STATE $time\_{periods} \longleftarrow {time\_periods\_in\_a\_given\_dataset}$
\STATE $n \longleftarrow number\_{of}\_{time}\_{periods}$
\STATE $ensemble\_{size} \longleftarrow n/2$ 
\WHILE{$k$ in $time\_{periods}$}
\STATE Train learner $L\textsubscript{i}$ on $k\textsubscript{i}$
\STATE $candidate \longleftarrow MeanSquaredError(L\textsubscript{i-1}, k\textsubscript{i})$
\IF  {$no\_{of}\_{learners}\_{in}\_{ensemble} \leq ensemble\_{size}$} 
\STATE Add $L\textsubscript{i-1}$ to $E$
\ELSE
\WHILE {$e$ in $E$}
\IF {$candidate <  MeanSquaredError(e\textsubscript{j},k\textsubscript{i})$}
\STATE replace $e\textsubscript{j}$ with $candidate$
\ENDIF
\ENDWHILE
\ENDIF
\ENDWHILE

\end{algorithmic}
\label{algo:sea}
\end{algorithm}

To understand how the temporal nature of AIOps data affects the generalizability of the derived interpretation of models updated with different approaches, we first divide the studied datasets into multiple time periods. We divide the Google dataset into 28 one-day time periods and the Backblaze dataset into 36 one-month time periods as we outline in Section~\ref{sec:data_preprocessing}. We then simulate the event of updating the constructed AIOps models to predict the outcome when the last time period becomes available (i.e., the 28th and 36th time period of Google and Backblaze datasets respectively). Unlike previous RQs, here we consider the available data from all the time periods (prior to the last time period) to update the AIOps model. We consider four approaches for updating the AIOps models including two periodic model retraining approaches (i.e., Sliding Window approach and Full History approach) and two time-based ensemble approaches (i.e., Accuracy Weighted Ensemble (AWE) and Streaming Ensemble Algorithm (SEA)). Finally, we compare if the derived interpretation of the AIOps models updated with these studied approaches capture the historic trends present in the data through the following three steps: 1) Test model generation, 2) Ground truth extraction, and 3) Interpretation comparison. 

\begin{figure}[!h]
    \centering
    \includegraphics[width=\textwidth]{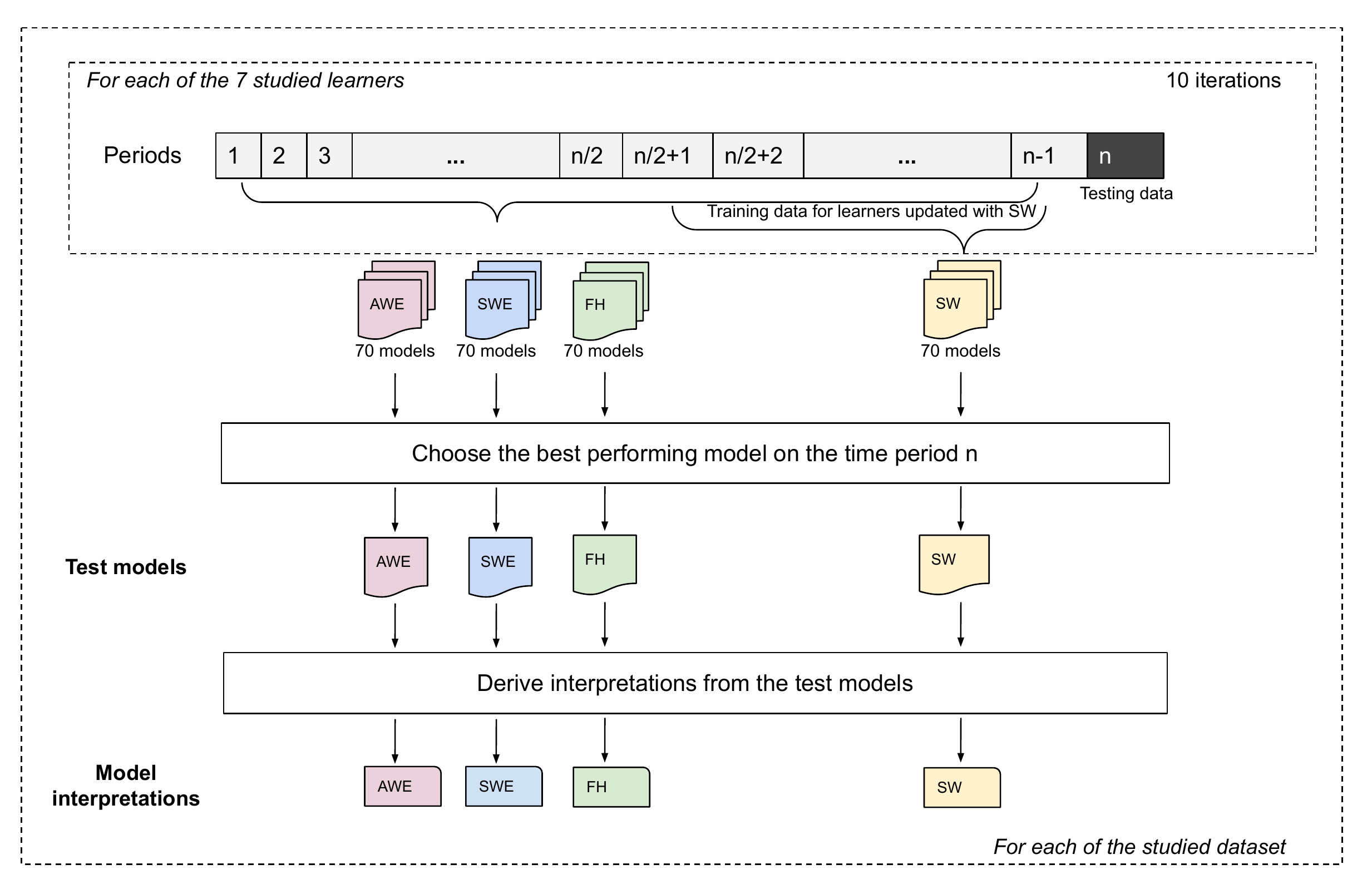}
    \caption{Overview of Step 1: Test model generation. FH stands for Full History, SEA stands for Streaming Ensemble Algorithm, AWE stands for Accuracy Weighted Ensemble, and SW stands for Sliding Window.
}
    \label{fig:rq3_flowchart_step_1}
\end{figure}

\smallskip\noindent\textbf{Step 1: Test model generation.} In this step, we first construct the AIOps models (whose interpretation we test) using data points from all the time periods (except the last one) to predict the outcome of data points on the last time period of the studied datasets. Figure~\ref{fig:rq3_flowchart_step_1} presents an overview of our test model generation step. We construct these AIOps models with all the 7 learners we outline in Section~\ref{sec:learners} and update them using the four model updating approaches. For each learner, we update the associated models with one of the studied model updating approach, the process is repeated for 10 iterations (with bootstrap sampling), resulting in 70 models being constructed. We do so across both the studied datasets. Among the 70 generated models for each model updating approach, we choose the best performing model (i.e., the model with the highest AUC on the last time period of each of the studied dataset a.k.a test dataset). We then derive the interpretation of the chosen best performing model (i.e., the test model). At the end, we obtain four derived interpretations from four test models, each from one of the four studied model updating approaches. We describe how each of the studied model updating approach leverages the historic data and updates the model below.

\smallskip\noindent\textbf{Periodic model retraining approaches.} These approaches typically retrain the model whenever data from a new time period becomes available. In our case study, we update the studied AIOps models to predict on the 28th and 36th time period of Google and Backblaze datasets respectively. 
\begin{itemize}
    \item \textit{Sliding window approach (SW)}: When updating the AIOps model with a SW approach, we retrain the model with n/2 time periods (i.e., half of the available time periods). For instance, when updating an AIOps model to predict the outcome on 28\textsuperscript{th} time period in Google dataset, we use the last 14 time periods to retrain the model. 
    
    \item \textit{Full history approach (FH)}: We use all the data available to retrain the AIOps model. For instance, when retraining an AIOps model to predict the outcome on 28\textsuperscript{th} time period in Google dataset, we use all the last 27 time periods. 
    
\end{itemize}


\smallskip\noindent\textbf{The time-based ensemble approaches.} Instead of updating a single AIOps model, the time-based ensembles combines several AIOps models trained on short periods of time (e.g., models trained on each time period (local models)) in an ensemble~\cite{streetKDD2001,wangKDD2003}. We then use this ensemble model to predict the outcome on future instances. For both the ensemble approaches that we use in this study, we set the ensemble size (i.e., the number of local models that we ensemble together) at n/2 similar to the SW approach. Below we describe the two ensemble approaches that we consider in our study.

\begin{itemize}
    \item \textit{Streaming Ensemble Algorithm (SEA)} combines multiple local models using a majority vote. Algorithm~\ref{algo:sea} presents the working of the SEA. To update the ensemble, SEA replaces the weakest model in the ensemble by observing which of the local models perform the worst in the latest time period for which the SEA is updated. The SEA ensembles multiple learners as follows:  When data points from a time period \textit{k} becomes available, a new learner \textit{L}\textsubscript{i} (e.g., RF learner) is trained on these data points to build a local model (please see line 7 in Algorithm~\ref{algo:sea}). Then, these data points are used to evaluate the model \textit{L}\textsubscript{i-1} trained on the previous time period \textit{k}\textsubscript{i-1} (please see line 8 in Algorithm~\ref{algo:sea}). If the number of models in the ensemble hasn't reached the predetermined ensemble size (n/2 in our case), the model \textit{L}\textsubscript{i-1} is simply appended to the ensemble (please see lines 9-10 in Algorithm~\ref{algo:sea}). Else, the Mean Squared Error (MSE) of the model \textit{L}\textsubscript{i-1} is compared to all the models in the ensemble. If the MSE of the model \textit{L}\textsubscript{i-1} is lower than that of all the models in the ensemble, the model \textit{L}\textsubscript{i-1} replaces the weak learner, if not, the model \textit{L}\textsubscript{i-1} is discarded (please see lines 10-16 in Algorithm~\ref{algo:sea}). When the next new set of data points become available for training, the model \textit{L}\textsubscript{i} pertaining to the current time period becomes the one that is tested against the other models in the ensemble. 
    
    \item \textit{Accuracy weighted ensemble (AWE)} is similar to the SEA except on two key points. First, instead of using a majority vote, AWE assigned weights to each model in the ensemble (we detail the weight assignment below). Second, when data from a new time period \textit{k} becomes available for training, SEA determines if model trained in the previous time period (\textit{L}\textsubscript{i-1}) should be kept or removed from the ensemble. Whereas, in AWE, we evaluate if the model (\textit{L}\textsubscript{i}) should be kept or discarded. 

\end{itemize}

 The weight for each model in the ensemble is assigned by calculating its prediction error on the latest training data time period \textit{k}\textsubscript{i}. More specifically, we calculate the Mean Square Error (MSE) of each model in the ensemble on the latest training data time period \textit{k}\textsubscript{i}. The MSE for each learner is calculated by computing the difference between the predicted probability of the given model and the observed outcome class (either 0 or 1) in \textit{k}\textsubscript{i}. We note it as MSE\textsubscript{i}. The weight for each model in the ensemble is then given by \textit{w}\textsubscript{i}=MSE\textsubscript{r}-MSE\textsubscript{i}, where MSE\textsubscript{r} is the MSE value of a model that predicts randomly (i.e.,MSE\textsubscript{r}=0.25). The models in the ensemble that perform worse than the MSE\textsubscript{r} are removed from the ensemble. We use a 10-fold cross validation to estimate the MSE.

\smallskip\noindent

\begin{figure}[!h]
    \centering
    \includegraphics[width=\textwidth]{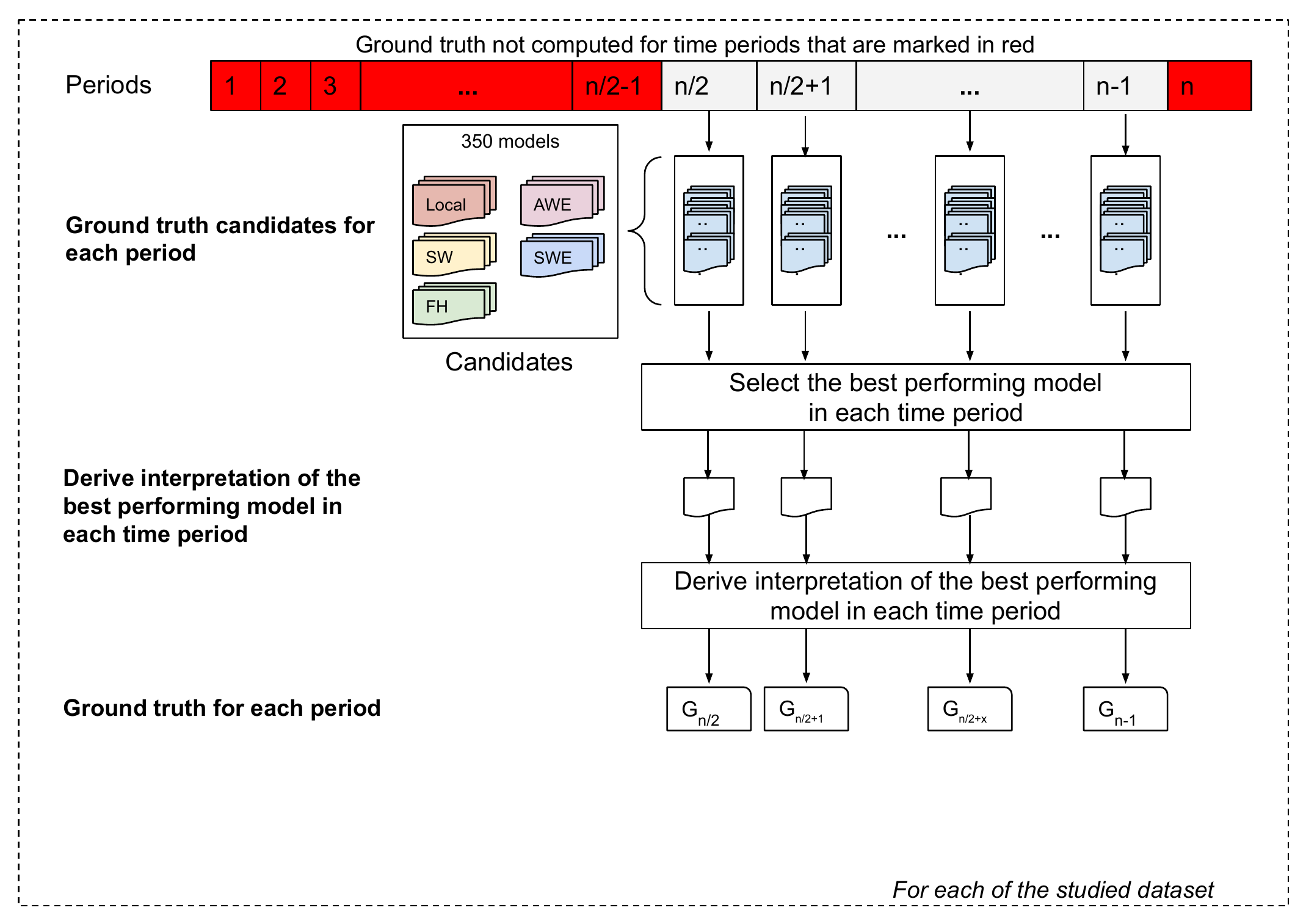}
    \caption{Overview of Step 2: Ground truth extraction. FH stands for Full History, SEA stands for Streaming Ensemble Algorithm, AWE stands for Accuracy Weighted Ensemble, and SW stands for Sliding Window.
}
    \label{fig:rq3_flowchart_step_2}
\end{figure}
\smallskip\noindent\textbf{Step 2: Ground truth extraction.} In this step, we extract the actual trends (i.e., ground truth) encapsulated in each time period of the studied datasets. Figure~\ref{fig:rq3_flowchart_step_2} presents an overview of our ground truth extraction step. We do so to verify if the derived interpretation of the test models obtained from the previous step are faithful to the historic data. While it is impossible to accurately know what features were the key drivers in prior time periods, as RQ2 results suggest, the interpretation of models in the high performing clusters are generally consistent. Furthermore, several prior studies also suggest that high performing models can typically be interpreted reliably~\cite{liptonQUEUE2018,rajbahadurTSE2020}. Therefore, to approximate the ground truth of each time period, we pick the best performing model in that time period and obtain its derived interpretation. For instance, to extract the ground truth of a given time period \textit{n-1}, we pick the model that is trained on \textit{n-1}\textsuperscript{th} time period (or on all the time periods before it) and has the best performance on the \textit{n}\textsuperscript{th} time period. We then derive the interpretation of this best performing model as a proxy for the ground truth contained in the time period \textit{n-1}. For example, in Google dataset, if we were to extract the ground truth for time period 25, then we consider the local models trained on 25\textsuperscript{th} time period and models trained on 25\textsuperscript{th} time period that are updated with the studied AIOps model updating approaches. Among these constructed models if we were to find that the RF model updated with FH approach to be the best performing model on the 26\textsuperscript{th} time period, we then derive the interpretation of this model. We consider this derived interpretation as the ground truth of the 25\textsuperscript{th} time period. 

We compute the ground truth from n/2 to n-1 (where n=\#time periods) time periods for each of the studied datasets by extracting the derived interpretation of the best performing model in the given time period. We do not compute the ground truth on the n\textsuperscript{th} time period as we are only trying to observe how faithful the best performing model on n\textsuperscript{th} time period is to the ground truth available in time periods n/2 to n-1. Furthermore, instead of computing the ground truth on all the time periods starting from 1, we do only from (n/2)\textsuperscript{th} time period, as we include both the AIOps models trained only on one time period (similar to RQ2) and AIOps models trained using the studied model updating approaches. However, the SW approach requires 1 to n/2 time periods to train. For instance, in Google dataset, for a learner to be updated with SW approach, the first 13 time periods are required to update the model trained on 14\textsuperscript{th} time period. We include both the models trained only on one time period and models trained using the studied model updating approaches since we want to maximize the chance of finding the best performing model in each time period. 

We train all the studied learners that we outline in Section~\ref{sec:learners} on time periods n/2 to n-1 of the studied datasets. We first train them using the same approach that we outline in RQ2. On each of the time period between n/2 and n-1 we first end up with 70 models. Next, we train our studied learners with the four model updating approaches first using time periods 1 to n/2 as the training data to predict on the outcome on (n/2+1)\textsuperscript{th} time period. We incrementally add each time period from (n/2+1)\textsuperscript{th} until n-1\textsuperscript{th} time period to the training data. For instance, in the Google dataset, we first build AIOps models on the 14\textsuperscript{th} time period, then we incrementally train models from the 14\textsuperscript{th} to the 27\textsuperscript{th} time period. Therefore, on each time period we end up with 280 models. Finally, among these $280 + 70 = 350$ ground truth candidate models constructed for each time period between n/2 to n-1 we compute the best performing model and interpret it. We consider derived interpretation of the best performing model in each time period as ground truth for the given time period. We do so because it best captures the variation of the dataset during that period. 

\begin{figure}[!h]
    \centering
    \includegraphics[width=\textwidth]{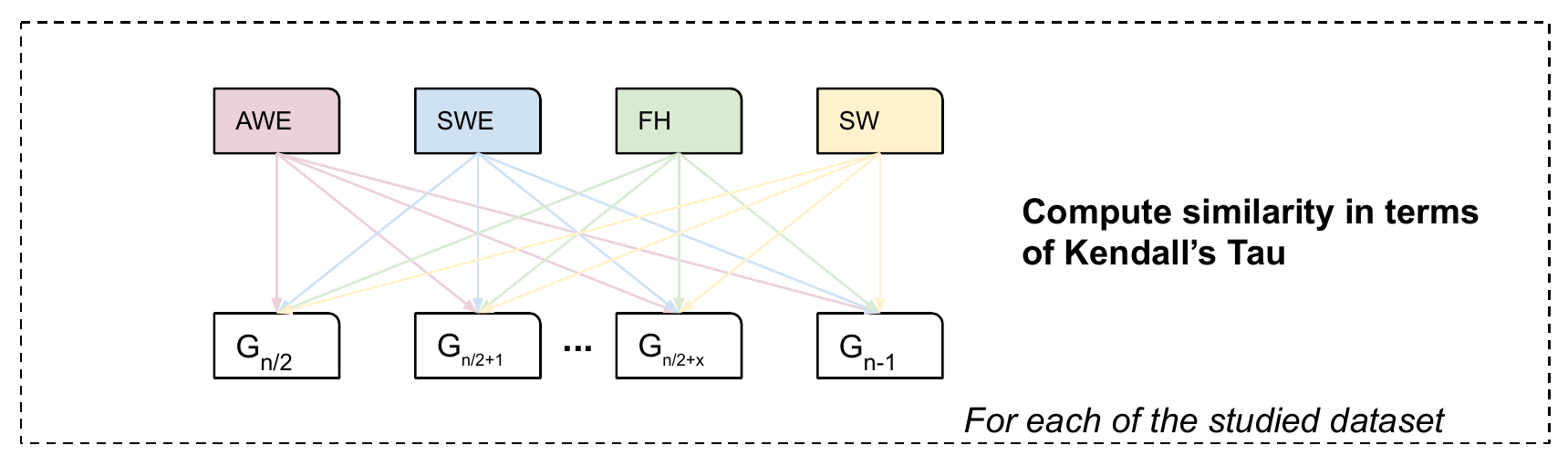}
    \caption{Overview of Step 3: Test model generation. FH stands for Full History, SEA stands for Streaming Ensemble Algorithm, AWE stands for Accuracy Weighted Ensemble, and SW stands for Sliding Window.
}
    \label{fig:rq3_flowchart_step_3}
\end{figure}
\smallskip\noindent\textbf{Step 3: Interpretation comparison.} To observe if the derived interpretations of the test models are impacted by the temporal nature of the AIOps data we compare its derived interpretation with the ground truth extracted in each prior time period. Figure~\ref{fig:rq3_flowchart_step_3} presents an overview of our interpretation comparison step. We do so by computing the similarity between the derived interpretation of test models (please note that four test models given by each studied model updating approach) and ground truth of each historic time period in terms of Kendall's Tau. We assert that temporal nature of the AIOps data does not impact the derived interpretation of test models, if the similarity scores given by Kendall's Tau is consistently strong across all historic time periods (i.e., n/2\textsuperscript{th} to n-1\textsuperscript{th} time periods).

Next, we seek to observe if the derived interpretation of a test model exhibits higher similarity to the ground truth across all the historic time periods compared to the others. If it does, the model updating approach of the given test model can be thought of as being the most faithful to the trends present in the prior data. To compute that, we conduct a paired Kruskall-Wallis H-test~\cite{kruskal1952use} between the computed similarity scores of the test models across all the time periods. A p-value $\leq 0.05$ on the Kruskall-Wallis H-test indicates that at least one of test model produces interpretation which has consistently higher similarity with the ground truth across the historic time periods. For instance, best performing test model updated by SEA to predict the outcome on 28\textsuperscript{th} time period would have 14 similarity scores in the Google dataset. Similarly the best performing test model updated by the other studied model updating approaches would also 14 similarity scores associated with each of them. We would then compute a Kruskall-Wallis H-test between these four similarity score distributions to see if the distribution associated with any one of the studied approaches is higher than the others.

If the Kurskall-Wallis test indicates that at least one of the test model has a different similarity score distribution than others on a given dataset, we also conduct a pairwise Wilcoxon-rank sum test and Cliff’s delta effect size test similar to RQ2. We do so in order to identify the best model updating approach(es) that yield(s) a significantly superior similarity with the ground truth in comparison to the other studied model updating approaches.

\subsection{Results.}

\begin{figure}
    \centering
    \includegraphics[width=\textwidth]{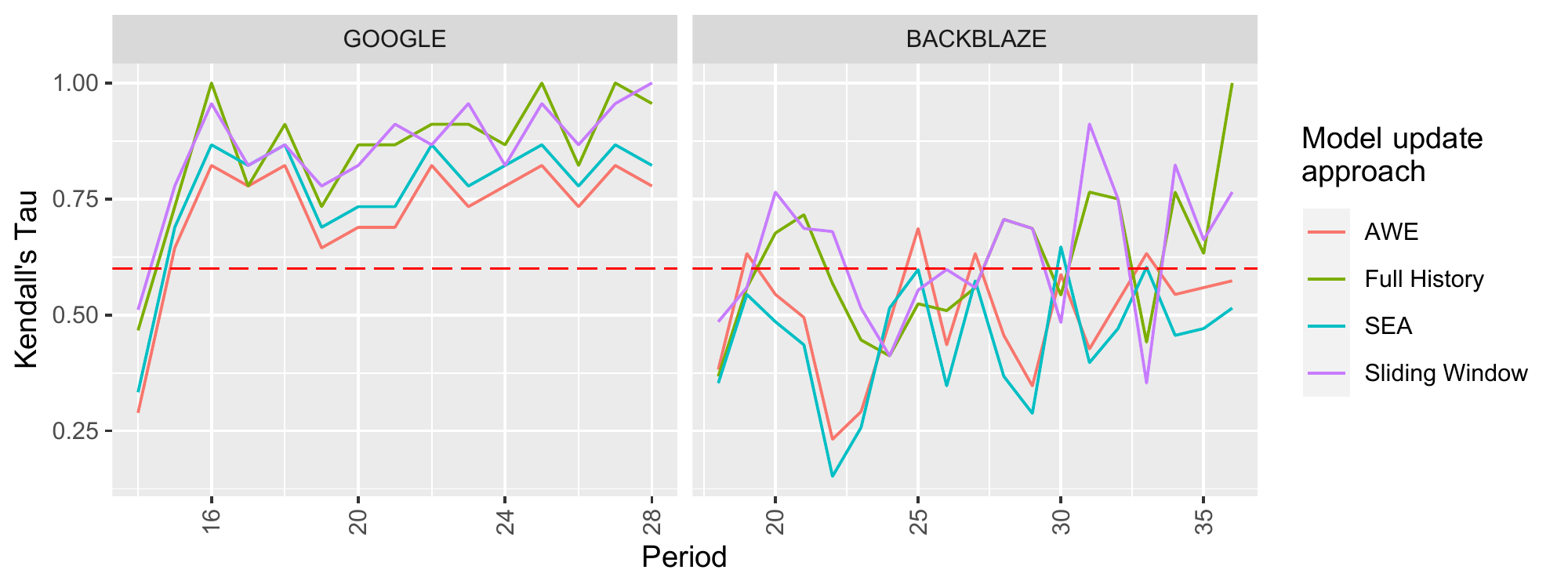}
    \caption{Similarity measurement (Kendall’s Tau) between the interpretation of studied AIOps modelling approaches and estimated ground truth. The red dashed line marks a Kendall's Tau of 0.6 (a strong agreement).}
    \label{fig:rq3_agreement}
\end{figure}

\begin{table}
\caption{Comparing the difference in similarity measurements (i.e., the similarity between the ground truth interpretation and derived interpretation of a AIOps model updated with the given update approach) between the derived interpretation for AIOps models updated with SW approach and the AIOps models updated with the other studied approaches.}
\label{tab:rq3_similarity_1}
\resizebox{\textwidth}{!}{
\begin{threeparttable}
\begin{tabular}{@{}llrrrrrr@{}}
\toprule
\multirow{3.5}{*}{\textbf{Update Approach}} & \multirow{3.5}{*}{\textbf{Dataset}} & \multicolumn{2}{c}{\textbf{FH}} & \multicolumn{2}{c}{\textbf{SEA}}      & \multicolumn{2}{c}{\textbf{AWE}}      \\ \cmidrule(l){3-8} 
& \multicolumn{1}{c}{}                                  & \textbf{\begin{tabular}[c]{@{}r@{}}Corrected\\p-value$^+$\end{tabular}}   & \textbf{Cliff's $d^*$}   & \textbf{\begin{tabular}[c]{@{}r@{}}Corrected\\p-value$^+$\end{tabular}} & \textbf{Cliff's  $d^*$} & \textbf{\begin{tabular}[c]{@{}r@{}}Corrected\\p-value$^+$\end{tabular}} & \textbf{Cliff's  $d^*$} \\ \midrule
\multirow{2.5}{*}{\textbf{SW}} & \textbf{Google}   & 0.86    & - (N/A)   & < 0.01           & 0.47 (L)           & < 0.01            & 0.69 (L)           \\ \cmidrule(l){2-8} 
& \textbf{BackBlaze}     & 0.35      &   - (N/A)   & < 0.01    & 0.49 (L)    & 0.02   & 0.64 (L)     \\ \bottomrule 
\end{tabular}
\begin{tablenotes}
\small
\item $^+$ $p < 0.05$ - Significant; $p \geq 0.05$ - Insignificant.
\item $^*$ N/A - Not Applicable; N - Negligible; S - Small; M - Medium; L- Large.
\end{tablenotes}
\end{threeparttable}}
\end{table}

\begin{table}
\caption{Comparing the difference in similarity measurements (i.e., the similarity between the ground truth interpretation and derived interpretation of a AIOps model updated with the given update approach) between the derived interpretation for AIOps models updated with FH approach and the AIOps models updated with the other studied approaches.}
\label{tab:rq3_similarity_2}
\resizebox{\textwidth}{!}{
\begin{threeparttable}
\begin{tabular}{@{}llrrrrrr@{}}
\toprule
\multirow{3.5}{*}{\textbf{Update Approach}} & \multirow{3.5}{*}{\textbf{Dataset}} & \multicolumn{2}{c}{\textbf{SW}} & \multicolumn{2}{c}{\textbf{SEA}}      & \multicolumn{2}{c}{\textbf{AWE}}      \\ \cmidrule(l){3-8} 
& \multicolumn{1}{c}{}                                  & \textbf{\begin{tabular}[c]{@{}r@{}}Corrected\\p-value$^+$\end{tabular}}   & \textbf{Cliff's $d^*$}   & \textbf{\begin{tabular}[c]{@{}r@{}}Corrected\\p-value$^+$\end{tabular}} & \textbf{Cliff's  $d^*$} & \textbf{\begin{tabular}[c]{@{}r@{}}Corrected\\p-value$^+$\end{tabular}} & \textbf{Cliff's  $d^*$} \\ \midrule
\multirow{2.5}{*}{\textbf{FH}} & \textbf{Google}   & 2.19    & - (N/A) & 0.001    & 0.47 (L)    & 0.001      & 0.63 (L)       \\ \cmidrule(l){2-8} 
& \textbf{BackBlaze}  & 2.68   & - (N/A)  & 0.006   & 0.57 (L)     & 0.03     & 0.40 (L)           \\ \bottomrule
\end{tabular}
\begin{tablenotes}
\small
\item $^+$ $p < 0.05$ - Significant; $p \geq 0.05$ - Insignificant.
\item $^*$ N/A - Not Applicable; N - Negligible; S - Small; M - Medium; L- Large.
\end{tablenotes}
\end{threeparttable}}
\end{table}

\textbf{The temporal nature of the AIOps data impact the similarity between the ground truth and the derived interpretation of the AIOps models updated using the studied model updating approaches.} From Figure~\ref{fig:rq3_agreement}, we observe that similarity between the ground truth and derived interpretation of the test models fluctuates across all the time periods for both studied datasets. In particular, except for 3 instances, none of the AIOps models constructed with the studied approaches have a similarity score of 1 (i.e., a perfect reflection of the ground truth) with the ground truth observed in the studied periods across both studied datasets. 

\textbf{The derived interpretation of AIOps models updated with SW approach and FH approach have a consistently higher similarity scores with the ground truth across all historic time periods than models updated with other studied approaches.} Figure~\ref{fig:rq3_agreement} shows that across the studied time periods, derived interpretation of test models trained with SW and FH approaches have a higher similarity score with the ground truth than the two time-based ensembles. In addition, we also observe that on Backblaze dataset, test models updated with AWE and SEA approaches have a strong similarity with ground truth only on 4 out 18 and 1 out of 18 time periods respectively. 

The Kruskall-Wallis test on the computed agreements between the interpretations of the test models and the ground truth resulted in significant (p<0.05) across both datasets. Such a result indicates that at least the similarity scores between the derived interpretation of the test models updated with one the studied model updating approach is significantly different from the similarity measurements produced by the other studied approaches. 

Table~\ref{tab:rq3_similarity_1} and Tabel~\ref{tab:rq3_similarity_2} presents the p-values of a pairwise Wilcoxon-rank sum test (p-value corrected) between the similarity scores of test models updated with SW and FH approach against the computed similarity scores of test models updated with the other studied approaches. From Table~\ref{tab:rq3_similarity_1} and Tabel~\ref{tab:rq3_similarity_2}, we find that the derived interpretation of best performing test models trained with both the SW model and the FH model produces significantly higher similarity with the ground truth than the test models updated with time-based ensembles approaches across both the studied datasets. Furthermore, we find that the magnitude of effect sizes is consistently large. However, we do not observe significant differences between the similarity measurements of the SW and FH models across either of the studied datasets.

\vspace{0.5cm}
\begin{Summary}{Summary of RQ3}{}
The temporal nature of the AIOps data indeed impacts the similarity between the interpretation of the AIOps models and the ground truth. Among the commonly used approaches to update AIOps models, we find that the derived interpretations from AIOps models updated with SW and FH approaches are the closest 
to the ground truth across all time periods. 
\end{Summary}
\section{Discussions and Future Work}
\label{sec:discussion}

In this section, we first discuss potential alternative experiment setups that could be applicable to our study, and then highlight the gap between AIOps studies and the reproducibility guidelines in the machine learning community. We would like to emphasize that the goal of our study is not to propose an optimal and improved AIOps model, but rather to reveal the threats or pitfalls of prior work’s practices when interpreting AIOps models. Therefore we keep the setup of the study close to prior work. Nevertheless, the alternative experiment setups discussed below can be examined in future work to confirm the generalizability of our findings under different setups.

\subsection{Training setups}
\label{sec:discussion-training}

In this study, we make several experimental choices in the training setups that we use to train our AIOps models. More specifically, 

\begin{enumerate}
    \item We choose to standardize the independent metrics before training a model;
    \item We downsample the training dataset prior to training the model;
    \item We do not use any synthetic class rebalancing method like SMOTE.
\end{enumerate}

We do so for two reasons. First, as we mentioned earlier, we wish to keep our AIOps model training setup as close to prior work as possible. All of the aforementioned experimental choices were used in prior work in the field of AIOps. For instance, Lyu et al.~\cite{YingzheSuppmaterial19} standardized the independent metrics of their input data before building their AIOps models. Similarly, several prior studies in AIOps~\cite{BotezatuKDD16, MahdisoltaniATC17} tend to use a downsampling strategy similar to ours. 

Second, the experimental choices that we make to build our AIOps models are quite robust. Though we tried to keep the setup of our study as close to prior work as possible, we took care to ensure that the experimental choices that we make are not sub-optimal. As \citet{thomas1998variable} and \citet{bring1995variable} state, the derived interpretations of a model are not impacted by the standardization methods such as StandardScaler. Similarly, we do not use advanced class rebalancing methods SMOTE to rebalance our datasets. Several prior studies~\cite{tantithamthavorn2018TSE2, turhan2012dataset, tantithamthavornICSE2018} show that rebalancing the datasets with techniques like SMOTE shifts the distribution of training data and impacts the derived interpretations. Since our study focuses on evaluating the consistency of interpretations derived from AIOps models, we avoided experimental choices that may impact the derived interpretations of a model. However, we highly encourage future research to study the impact of different experimental choices on the consistency of the interpretation derived from AIOps models using the rigorous set of criteria that we outline in our study.

\subsection{Evaluation setups}
\label{sec:discussion-evaluation}

In this study, we used AUC as the metric to evaluate the performance of models. As stated in Section~\ref{sec:setup-evaluation}, we selected AUC as the evaluation metric because prior work~\cite{tantithamthavornICSE2018} has shown that threshold-independent metrics such as AUC provides more stable evaluation results than threshold-dependent metrics such as precision, recall and F-measure.

In particular, we used Area Under ROC Curve (AUROC) in our study. ROC curve, or Receiver Operating Characteristic curve, is a curve that describes the relationship between a model’s true positive rate and false positive rate~\cite{bewick2004statistics}. The AUROC value has a clear interpretation schema. An AUROC value of 0.5 indicates a model performance that is equivalent to random guessing, while an AUROC value of 1 indicates a model that can provide perfect prediction with 0 error rate under a certain threshold. However, when a given dataset is extremely imbalanced (like in many AIOps datasets), AUROC can potentially produce over-optimistic evaluation results. 

An alternative metric that is insensitive to imbalanced data is Area Under Precision-Recall Curve (AUPRC), which describes the relationship between a model’s precision and recall. However unlike AUROC which has a fixed baseline of 0.5, AUPRC does not have a fixed baseline for comparison. In contrast, the baseline to evaluate whether an AUPRC value is good depends on the ratio of positive samples in the dataset~\cite{boyd2012unachievable}. In RQ2, we studied the relationships between model performance and external consistency of the interpretation. To study such relationships, we compared and clustered many different models that are not necessarily trained on the same sample of the dataset. In cases such as those, it is not practical to use AUPRC in our study setup to conduct a fair comparison across models (since AUPRC does not have a fixed baseline for different datasets like AUROC). In addition, using AUPRC would not let us suggest a general acceptable threshold as a guideline for AIOps model interpretation.

\subsection{Interpretation levels}
\label{sec:discussion-interpretation}

In this study, we used a model-level, model-agnostic approach (i.e., the permutation feature importance) to interpret AIOps models, as prior work on AIOps mostly focus on model-level interpretation. However, in some cases, an instance-level interpretation could be useful  for improving the model quality and enhancing the trustworthiness of AIOps models. For example, it is possible that, although two instances have similar features, the prediction results differ (e.g., one predicted as a successful job run and the other predicted as a failed job run). In such cases, it is important to understand the rationale behind the predictions, which could provide insights on how to improve the performance of the AIOps model in the future. 

Recently, many instance-level interpretation techniques (e.g., LIME~\cite{RibeiroKDD16}, Anchor~\cite{ribeiro2018anchors}, and SHAP~\cite{lundberg2017unified}) have been proposed in the research literature~\cite{RibeiroKDD16, ribeiro2018anchors, lundberg2017unified, pedreschi2019meaningful, guo2018lemna}. The general idea of instance-level interpretation techniques is to first use local surrogate models (e.g., a linear regression model) to approximate the predictions made by complex models (e.g., a deep neural network model), and then derive interpretations from the local surrogate models which are inherently interpretable. There are two common types of derived interpretations for a prediction: the feature importance ranks and a set of rules and depending on the need both of these can be useful. For instance, Zhao et al.~\cite{ZhaoFSE20} applied LIME to generate a visualized report for AIOps engineers to understand each incident prediction result. 

While it is possible (and useful) to apply instance-level interpretation in AIOps models, a recent study~\cite{FanTIFS21} in the malware detection domain evaluated five instance-level interpretation techniques. They found that the derived interpretations are not consistent even if the prediction is made by the same model with the same input. Within the five studied instance-level interpretation techniques, LIME~\cite{RibeiroKDD16} achieves the most stable results when the model is slightly changed. Such findings indicate a need for a study similar to ours to provide guidelines to ensure consistent interpretations can be obtained from AIOps models. However, since our study only focused on model-level interpretation, it remains unknown whether our findings apply to instance-level interpretations in general and we suggest that future work should explore this in detail.  

\subsection{The gap between reproducibility guidelines in the machine learning community and AIOps studies}
\label{sec:discussion-gap}

There is a well-accepted reproducibility  checklist~\cite{pineau2020improving} proposed in the machine learning community which outlines the requirements when submitting research manuscripts to support better reproducibility of the important findings. The checklist contains 21 guidelines which are listed in the appendix of this paper. There are five categories of the 21 guidelines: models/algorithms, theoretical claims, datasets, shared code, and experimental results. In order to understand whether the recent AIOps studies follow the guidelines proposed in the machine learning community, we revisited the 11 previously surveyed AIOps studies in Section~\ref{sec:backgroud:issues}. 

All the studies that we included in our survey have followed the models/algorithms related guidelines presented in the reproducibility guideline. None of the surveyed studies propose any theoretical claims, hence guideline No. 4 and No. 5 do not apply. For the guidelines related to datasets, only 2 out of the 11 studies provided a downloadable link for the dataset that they use in their study. 6 out of the 11 studies did not explain the detailed process on how they collect the data if newly presented, against guideline No.10. For the guidelines related to shared code, 10 out of 11 studies fail to provide training code, evaluation code, or (pre-) trained models, which goes against guideline No. 12, 13, and 14. None of the surveyed studies provide precise commands in the README file to reproduce the results, which goes against guideline No. 15. For the experimental result part,  5 out of 11 studies did not mention the approaches that they use for selecting the hyperparameters. 7 out of 11 studies did not specify the exact number of training and evaluation runs.

Our results show that there is currently a gap between recent AIOps studies and the reproducibility guidelines in the machine learning community. Even though it is understandable that the guidelines are not explicitly followed (for example, for guideline No. 9, it is reasonable that certain datasets cannot be disclosed due to company policies, and the guidelines were only introduced in 2019), it is concerning to find that so many studies violate the guidelines for reproducibility. In addition, even by following the reproducibility guidelines, the results might not be reproducible due to uncontrolled randomness (as we show in RQ1). Hence, our first guideline regarding the internal consistency complements the reproducibility checklist. It is crucial for the future AIOps studies to start accounting for both the reproducibility guidelines and our guidelines to address reproducibility concerns and subsequently achieve the internal consistency of interpretations.
\section{Guidelines to Practitioners}
\label{sec:guideline}

In this section, based on our findings, we recommend the following practical guidelines for AIOps researchers and practitioners to reliably interpret their AIOps models

\smallskip\noindent\textbf{[Guideline 1]: \emph{Always find ways to expose and record the random seeds used to control the non-determinism from the learner, hyperparameter tuning, and data sampling when building an AIOps model.}} From the results presented in Section~\ref{sec:rq1}, we observe that identical and internally consistent interpretations could be derived from the constructed AIOps models only when common sources of non-determinism were controlled. It is important to ensure the internal consistency of the derived interpretations of an AIOps model to enable managers and DevOps engineers to act based on these interpretations. For instance, if a DevOps engineer gets a different feature as the source of a problem every time the an AIOps model is retrained (even on the same setup), then it would be impossible to act upon. More importantly, as~\citet{krishnaTSE2018} outline, the DevOps engineer would lose trust in the constructed AIOps model when obtained insights are not consistent.

\smallskip\noindent\textbf{[Guideline 2]: \emph{Only use AIOps models with a minimum acceptable performance (i.e., AUC greater than 0.75) to derive interpretations.}} From the results presented in Section~\ref{sec:rq2}, we note that derived interpretations of models with low performance (i.e., AUC less than 0.75) exhibit very low external consistency. Only when the model's performance in terms of AUC is greater than 0.75 the derived interpretations among similarly performing models start to become consistent. In light of these findings, we caution against the usage of lower performing interpretable models in lieu of higher performing ones for deriving interpretations~\cite{LiTOSEM20}. Particularly since, interpretations derived from such models may not accurately reflect the ground truth. 

\smallskip\noindent\textbf{[Guideline 3]: \emph{When using AIOps models to derive interpretations, to update them when new data becomes available, use periodic retraining strategies like Sliding Window and Full History approaches.}} As we mention in Section~\ref{sec:guideline}, it is common for AIOps researchers and practitioners to update their trained AIOps models to keep up with the constant evolution of the AIOps data. When doing so, as we observe from the results presented in Section~\ref{sec:rq3}, it is possible for the updated models to lose sight of the historic trends. Only the models updated with Sliding Window and Full History approaches consistently exhibit high similarity with the historic ground truth present in the data. 

\section{Threats to Validity}
\label{sec:threats}
In this section, we discuss the threats to validity of our case study.

\subsection{Internal Validity}

In this paper, we studied four different AIOps model update approaches and compared their interpretation. The window size of the sliding window approach is set to be a half of the number of total time period, which is consistent with prior work~\cite{LyuTOSEM21}.

\subsection{External Validity}

In this paper, we used seven learners on two datasets and leveraged the permutation feature scores for interpretation. The two datasets (the Google cluster trace dataset and the Backblaze hard drive statistics dataset) and the seven learners are widely adopted in previous AIOps studies~\cite{LyuTOSEM21,LiTOSEM20}. Although our results might not generalize when using other learners, datasets, and interpretation methods, our case study setup is generic and can be applied to other types of predictive AIOps tasks in future work.

For AIOps tasks that leverage other types of machine learning algorithms such as unsupervised learning, active learning, or reinforcement learning, the generalizability of our findings should be examined in future work. In addition, our study only focuses on AIOps tasks. Hence, generalizability of our guidelines should be examined in other domains.

In RQ2, we derive a conclusion that models with a minimum AUC value of 0.75 tend to have high consistency among their interpretations. This conclusion is based on the empirical observations made on the two studied datasets. For other datasets and tasks, it is possible that models that do not reach the minimum AUC value threshold (i.e., 0.75) could also have high consistency among their interpretations. Future work is needed to study the generalizability of our conclusions

We implemented our experiments using Python and the 0.24.0 version of scikit-learn API. We did not investigate the impact of different implementations in other programming languages (e.g., R, C++) and other popular machine learning frameworks such as Weka and tensorflow. 
In addition, we only used the permutation feature importance scores provided by the scikit-learn API to derive the interpretations from each model while there are other implementations~\cite{pimp,eli5}. Although we think this technique fits best in our context, there might be other techniques available to quantitatively derive the interpretations. Future study is encouraged to investigate this matter. 

\subsection{Construct Validity}

In the setting of all RQs, we downsample the majority class in the training set so that the success-to-fail ratio is 10:1, which may present a threat to our construct validity. However, this process is consistent with prior work to mitigate the imbalance of the dataset~\cite{BotezatuKDD16,MahdisoltaniATC17}.

In the setting of RQ1, we repeat the model training process for at least 10 times for each learner to observe the impact of randomness. Similarly, in RQ2, the performance threshold of producing consistent interpretation might be impacted by the characteristics of the randomly sampled datasets. We repeated the training task for each learner in each period of data for at least 10 times to increase the variety of sampled dataset and trained models, and mitigate the risk.

To measure the consistency of the interpretations, we used three types of similarity measurements: the Kendall's W, top 5 overlap score, and top 3 overlap score. These measurements have been widely used in previous studies\cite{LyuTOSEM21,rajbahadurTSE2020}. 

We applied an automated process (i.e., randomized search) for hyperparameter tuning in the training process and set the iteration times to be 100. We choose this method over manually tuning as it would be impractical to manually tune every model with randomly sampled data. In addition, Bergstra and Bengio ~\cite{BergstraJMLR12} show that randomized search is efficient in locating the optimal hyperparameters. Even though we cannot guarantee that we extract the best set of hyperparameters, the impact on our study results is limited, since our experiments are conducted on two datasets and seven learners, and our findings do not rely on the hyperparameters being the most optimal.

\section{Conclusions}
\label{sec:conclusion}

In this paper, we study the consistency of interpretation of AIOps models through a case study on two popular AIOps use cases: (1) job failure prediction using the Google cluster trace dataset; and (2) hard drive failure prediction using the Backblaze hard drive statistic dataset. We assess the consistency of AIOps model interpretation along three dimensions: internal consistency, external consistency, and time consistency. 
Our results show that inherent randomness from learners, randomized hyperparameter searching, and sampling randomness can lead to internally inconsistent interpretations of AIOps models. In addition, we observe that the performance of AIOps models impact the external consistency of interpretations---the interpretations of AIOps models in the highest-performing clusters tend to be more consistent than those of other models. Finally, we find that AIOps models built using Sliding Window and Full History approaches have the most consistent interpretations to the evolving ground truths.

In light of these findings, we suggest the AIOps researchers and practitioners to: 1) Always control the non-determinism from the learner, hyperparameter tuning, and data sampling by exposing and recording random seeds; 2) Use models with high performance (i.e., AUC greater than 0.75) to derive interpretations; and 3) Use either a Sliding Window or Full History approach to update the model when using the updated model to derive interpretations. 
Our findings and guidelines can help AIOps researchers and practitioners derive consistent interpretations from AIOps models. For the AIOps applications such as issue prediction and the issue mitigation, by following our guidelines, practitioners can derive more reliable and consistent interpretations that explain the occurrence of failures or outages in the field, which can help them make better managerial decisions that have long-lasting effects and develop better tooling support that saves maintenance efforts.

In the future, we plan to extend our study to more types of AIOps tasks such as performance anomaly detection and diagnosis and other types of interpretation techniques, such as instance-level interpretations. In addition, future work can compare the interpretation between manually tuned models and automated tuned models. We are also interested in applying quality assurance techniques such as metamorphic testing and differential testing to further verify the quality of the model interpretations within the AIOps context.  

\bibliographystyle{ACM-Reference-Format}
\bibliography{11-references}



\end{document}